\documentclass[letterpaper,journal]{IEEEtran}

\usepackage{amsfonts}
\usepackage{array}
\usepackage[caption=false,font=normalsize,labelfont=sf,textfont=sf]{subfig}
\usepackage{textcomp}
\usepackage{stfloats}
\usepackage{url}
\usepackage{verbatim}
\usepackage{graphicx}
\def\BibTeX{{\rm B\kern-.05em{\sc i\kern-.025em b}\kern-.08em
    T\kern-.1667em\lower.7ex\hbox{E}\kern-.125emX}}
\usepackage{balance}

\usepackage[utf8]{inputenc} 
\usepackage[T1]{fontenc}
\usepackage{url}
\usepackage{ifthen}
\usepackage{cite}
\usepackage[cmex10]{amsmath} 
\usepackage{makecell, booktabs} 
\usepackage{pifont}
\newcommand{\cmark}{\ding{51}} 
\newcommand{\xmark}{\ding{55}} 
\usepackage{tikz}
\usepackage{comment}
\usepackage{graphicx} 
\usepackage{amssymb}
\usepackage{amsthm}
\usepackage{hyperref}
\usepackage{algorithm}
\usepackage{algpseudocode}
\usepackage{xcolor}
\usepackage{tcolorbox}

\theoremstyle{definition}

\newtheorem{example}{Example}

\usepackage{algorithm}
\usepackage{algpseudocode}

\newcommand{\pare}{\eta}

\newcommand{\gdot}{g(.)}

\newcommand{\bu}{\mathbf{u}}
\newcommand{\by}{\mathbf{y}}
\newcommand{\bn}{\mathbf{n}}

\newcommand{\est}{\mathsf{est}}
\newcommand{\DC}{\mathsf{DC}}
\newcommand{\AD}{\mathsf{AD}}

\newcommand{\PA}{\mathsf{PA}}

\newcommand{\bestu}{\mathsf{U}^*_{a,b}}

\newcommand{\bestetaalgnew}{\hat{\eta}^*_{\delta,\lambda}}

\begin{document}

\title{Game of Coding: Coding Theory in the Presence of Rational Adversaries, Motivated by Decentralized Machine Learning
}
\author{%
  \IEEEauthorblockN{Hanzaleh Akbari Nodehi,}
  \and
  \IEEEauthorblockN{Viveck R. Cadambe,}
  \and
  \IEEEauthorblockN{Mohammad Ali Maddah-Ali}
}
 \maketitle

\begin{abstract}
Coding theory plays a crucial role in enabling reliable communication, storage, and computation. Classical approaches assume a \emph{worst-case adversarial} model and ensure error correction and data recovery only when the number of honest nodes exceeds the number of adversarial ones by some margin. However, in some emerging decentralized applications, particularly in decentralized machine learning (DeML), participating nodes are rewarded for accepted contributions. This incentive structure naturally gives rise to \emph{rational adversaries} who act strategically rather than behaving in purely malicious ways.

In this paper, we first motivate the need for coding in the presence of rational adversaries, particularly in the context of outsourced computation in decentralized systems. We contrast this need with existing approaches and highlight their limitations. We then introduce the \emph{game of coding}, a novel game-theoretic framework that extends coding theory to trust-minimized settings where honest nodes are not in the majority. Focusing on repetition coding, we highlight two key features of this framework: (1) the ability to achieve a non-zero probability of data recovery even when adversarial nodes are in the majority, and (2) Sybil resistance, i.e., the equilibrium remains unchanged even as the number of adversarial nodes increases. Finally, we explore scenarios in which the adversary’s strategy is unknown and outline several open problems for future research.

\end{abstract}
\section*{Introduction}
Coding theory plays a foundational role in ensuring the correctness and robustness of modern communication, computation, and storage systems. It provides systematic methods to protect against errors, system failures, and even adversarial interference. 
Its scope spans both exact recovery of discrete data over finite fields, and approximate recovery in analog settings~\cite{SudanBook, ZamirCoded, jahani2018codedsketch, roth2020analog, moradi2025general, dutta2019optimal, dutta2016short, yu2017polynomial, yu2020entangled, yu2019lagrange}.

\begin{figure}[t]
    \centering
    \includegraphics[width=0.35\textwidth]{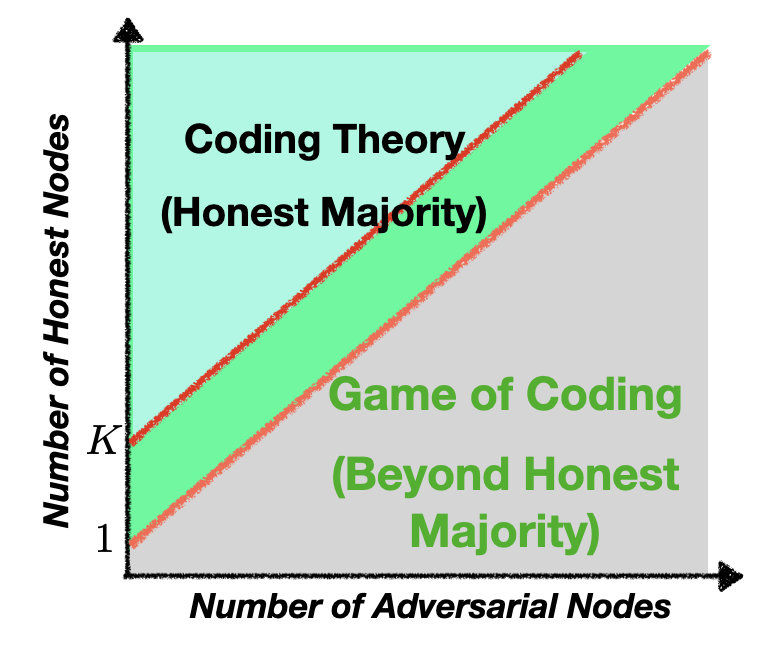}
    \caption{
    Game of coding aims to address scenarios that the trust assumptions of classical coding theory no longer hold, particularly in decentralized environments\cite{sliwinski2019blockchains}.
    }
    \label{fig:region}
\end{figure}

Despite its broad applicability in both discrete and analog regimes, coding theory fundamentally relies on \emph{trust assumptions} to enable reliable data recovery. Consider, for instance, the setting of distributed storage, where a data file is encoded and distributed across \( N \) nodes. 
Some storage nodes, referred to as \emph{honest} nodes, return exactly what was stored there in the first place. On the other hand, some storage nodes, either due to adversarial control or faults, may return corrupted data. 
If a maximum distance separable (MDS) code of dimension \( K \) and length \( N \) is used, then at least \( \lfloor (N-K)/2 \rfloor + 1 \) honest nodes are required for correct decoding. In the extreme case of \( K = 1 \) (i.e., repetition coding), this reduces to the requirement of an \emph{honest majority} (see Fig.~\ref{fig:region}). Similar conditions are necessary for analog coding as well~\cite{roth2020analog}. This same principle has also been extended to apply coding techniques for computation, to improve the reliability of distributed computing as well.

In this work, we focus on a trust-minimized regime where honest nodes are not necessarily in the majority (see Fig. \ref{fig:region}). Under conventional coding-theoretic models, the absence of these guarantees allows an adversary to manipulate the data collector (DC) into rejecting all inputs, resulting in a denial-of-service attack with no output. However, we show that by replacing this overly conservative worst-case adversary model with a rational adversary, one that seeks to maximize its own utility rather than cause arbitrary disruption, reliable estimation remains achievable. 

This shift is motivated by emerging decentralized applications, where trust assumptions are often unverifiable or explicitly violated\cite{sliwinski2019blockchains}. In such environments, the adversary is subject to an incentive-oriented structure, typically being rewarded for accepted contributions and even sometimes penalized for rejection \cite{Buterin_Slashing}, which naturally steers adversaries towards rational rather than worst-case behavior. Indeed, \emph{rational} adversarial behaviour is foundational in real-world design of decentralized systems (see, for example Ethereum \cite{Buterin_Slashing}.) 

This tutorial outlines a new framework, called the \emph{game of coding}, that extends classical coding theory to account for rational adversarial behavior. The framework is motivated by the emerging application of decentralized machine learning (DeML). This tutorial (1) presents the core vision of DeML, its main challenges, and how computation outsourcing can help address them; (2) reviews existing computation outsourcing approaches, with a focus on DeML, and identifies their limitations; (3) develops the game of coding framework; and (4) presents key results derived from its analysis. We conclude with a discussion of open research directions to further advance the field.

\section*{Motivation: Decentralized Machine Learning (DeML)}
\subsection*{Decentralization Movement}
The emergence of Bitcoin~\cite{bitcoin2008bitcoin} marked a pivotal moment in computing, demonstrating that secure, verifiable financial transactions could occur without reliance on a centralized entity. 
Building on this foundation, decentralized computing platforms such as Ethereum~\cite{buterin2013ethereum} extended the model from digital currency to general-purpose computation. 
As a result, they have enabled a wide range of applications, including decentralized finance, supply chain tracking, digital identity, and asset tokenization~\cite{ruoti2019sok}.

\begin{figure}[t]
    \centering
    \includegraphics[width=0.40\textwidth]{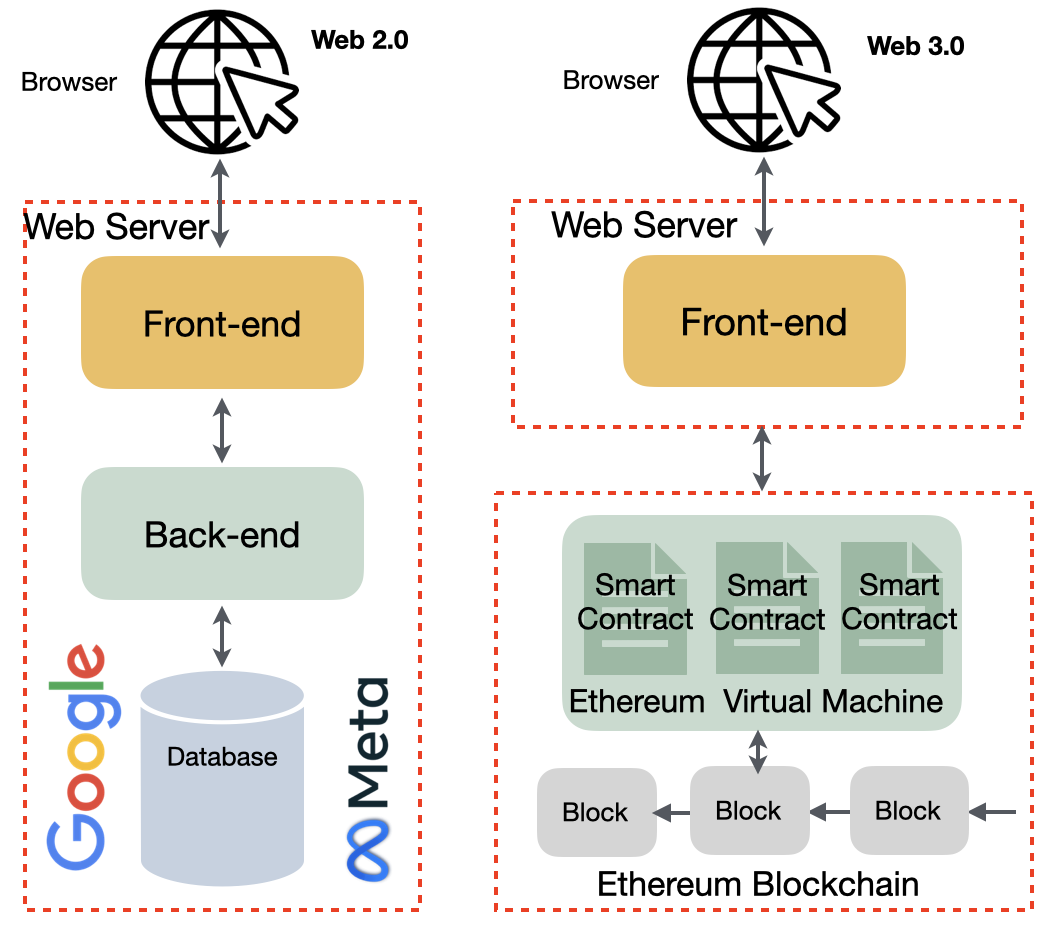}
    \caption{
    In Web 2.0, back-ends rely on powerful corporate servers with limited transparency. Web 3.0 uses decentralized platforms, which are transparent but computationally limited, requiring heavy tasks to be outsourced without an honest majority guarantee.
    }
    \label{fig:web3}
\end{figure}

This success has fueled the broader vision of Web 3.0, an open, decentralized alternative to the current Web 2.0 ecosystem (see Fig. \ref{fig:web3}). While Web 2.0 infrastructure is dominated by centralized service providers with opaque data and compute layers, Web 3.0 aspires to rebuild these layers on decentralized, transparent and permissionless networks, fostering greater fairness, resilience, and censorship resistance.

\subsection*{The Main Challenge of DeML}
A particularly exciting application of Web 3.0 is decentralized machine learning (DeML), which has recently attracted growing attention and led to the emergence of numerous start-ups (e.g., \textsc{Sentient}, \textsc{FedML}, \textsc{Gensyn}, \textsc{Kosen Labs}, \textsc{EZKL}, \textsc{Together}, \textsc{Talus}, \textsc{Olas}, \textsc{Bagel}, \textsc{Allora}, \textsc{Sahara AI}, \textsc{Ritual}, \textsc{Theoriq}). 

In Web 2.0 AI systems, ML workloads are typically executed on servers operated by providers like Google or Amazon, with limited transparency. In contrast, DeML aims to use blockchains as a shared execution and coordination layer, where ideally every step of model training or inference can be verified directly on the blockchain (i.e., on-chain), enabling secure, live, and accountable AI.

However, blockchains are not well-suited for heavy computational workloads. They are inherently slow and resource-constrained. This limitation stems from the overhead introduced by consensus algorithms, which effectively reduce the system’s net computing power to that of a standard desktop computer.

To address this limitation, one common approach is to outsource the computation to external volunteer nodes.  In fact, this idea forms the core of blockchain scalability solutions called Rollups\cite{ kalodner2018arbitrum}, which at the time of writing, more than $40$ billion in assets are secured using them; the curious reader is encouraged to visit the link at reference \cite{l2beat}. The problem, however, is that these nodes are not always trustworthy.  Their presence undermines the reliability of the results and poses a major challenge for safe and scalable computation outsourcing.

\subsection*{Computation Outsourcing Requirements}
 Our goal is to design a secure outsourcing framework that remains reliable even in the presence of adversarial nodes. A fundamental requirement is that the verification process must be lightweight and efficient. Beyond that, the approach is guided by four key requirements:

\begin{enumerate}
\item \textbf{Low computational overhead:} The outsourcing framework should avoid increasing the computational burden on those  external nodes. This is essential to guarantee that the system remains open and inclusive, avoiding a reliance on nodes with disproportionately high computational resources. Otherwise, participation would be limited to a small set of powerful actors, undermining decentralization and introducing central points of control. Enabling more nodes to participate not only strengthens the system’s resilience, by reducing the risk of concentrated failure or collusion, but also promotes a more democratized form of machine learning, where diverse and widely distributed contributors can engage in the computation process.

\item \textbf{Support for Approximate Computing:} Many practical workloads involve computations that are not exact and do not take place over finite fields. Instead, they rely on real-valued and approximate operations such as quantization, rounding, fixed-point or floating-point arithmetic, randomized sketching, sampling, and random projection. A robust outsourcing framework should natively support such approximate computations without requiring extensive reformulation. This capability is particularly critical in machine learning applications, where approximate methods are widespread.

\item \textbf{Fast Finality:} The system should confirm results quickly and decisively. In some existing approaches, results remain undecided for an extended period after submission, during which they can still be invalidated or changed. This prolonged uncertainty slows down dependent processes. A practical computation outsourcing framework should avoid such delays and ensure that results are finalized without long waiting times.

\item \textbf{Resilience in adversarial-majority settings:} The outsourcing framework must remain effective even when the majority of external nodes are adversarial. This is crucial in decentralized environments where trust is a scarce resource, and assuming an honest majority is often unrealistic \cite{sliwinski2019blockchains}. Designing with this in mind ensures robustness even under pessimistic trust assumptions. 
\end{enumerate}

\section*{A Review of Solutions for Computation Outsourcing}\label{sec:related_works}
 
In this section, we review state-of-the-art approaches to computation outsourcing, particularly in the context of DeML applications, and highlight their key limitations (see Table~\ref{tab:comparison} for a high-level summary).

\begin{table*}[t]
\centering
\caption{Comparison of existing approaches for computation outsourcing, particularly in the context of DeML applications, with the proposed solution, namely the game of coding framework.}

\label{tab:comparison}
\resizebox{\textwidth}{!}{%
\begin{tabular}{@{}lcccc@{}}
\toprule
\textbf{Method} & 
\makecell{\textbf{Low Computational}\\\textbf{Overhead}} & 
\makecell{\textbf{Support for }\\\textbf{Approximate Computing}} & 
\makecell{\textbf{Fast Finality }} & 
\makecell{\textbf{Resilience in}\\\textbf{Adversarial-Majority Settings}} \\
\midrule
Verifiable Computing \cite{feng2021zen, ghodsi2017safetynets, zhao2021veriml, liu2021zkcnn, garg2023experimenting}        & \xmark & \xmark & \cmark & \cmark \\
Optimistic Verification \cite{bhat2023sakshi, conway2024opml}                & \cmark & \xmark & \xmark & \cmark \\
Coded (Redundant) Computing \cite{jahani2018codedsketch, moradi2025general, roth2020analog}   & \cmark & \cmark & \cmark & \xmark \\
Game of Coding \cite{GOC_Firstpaper, GoDSybil, GOC_unknown}              & \cmark & \cmark & \cmark & \cmark \\
\bottomrule
\end{tabular}
}
\end{table*}

\subsection*{Verifiable Computing}

\begin{figure}[t]
    \centering
    \includegraphics[width=0.45\textwidth]{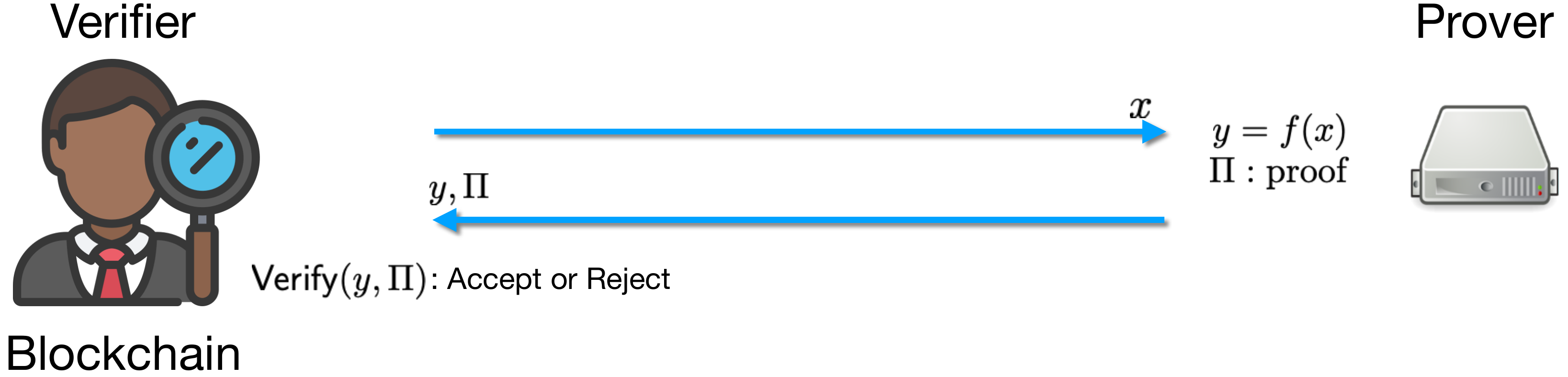}
    \caption{ In verifiable computing, the prover performs the computation off-chain and sends both the result and a validity proof to the verifier (typically a smart contract on the blockchain). The verifier checks the proof and either accepts or rejects the output without re-executing the computation. This architecture provides \emph{fast finality} and \emph{resilience in adversarial-majority settings}. However, it incurs high computational cost for proof generation and lacks compatibility with approximate, real-valued workloads.
}

    \label{fig:gameversusverifiable}
\end{figure}

One widely studied approach for secure computation outsourcing is \emph{verifiable computing}~\cite{thaler2022proofs}. Verifiable computing builds on cryptographic techniques that allows users to verify the correctness of a computation without re-executing it in full. This paradigm has a long lineage in theoretical computer science and cryptography, and has recently seen renewed interest in practical settings such as cloud computing and decentralized blockchain-based applications.
In this model, a heavy computation \( y = f(x) \) is offloaded to external computing nodes. These nodes play the role of the \emph{prover}. They receive an input \( x \), and compute an output \( y = f(x) \), where we assume that evaluating the function \( f(x) \) is too resource-intensive to be performed on-chain. As a result, the task is outsourced to the prover. Alongside the output, the prover generates a cryptographic artifact \( \Pi \), known as a \emph{validity proof}, which certifies the correctness of the result. The prover returns both \( y \) and \( \Pi \) to the \emph{verifier}. The verifier, typically implemented as a program referred to as a smart contract, runs on a blockchain network and acts as the DC, applies a verification function \( \mathsf{Verify}(y, \Pi) \) to decide whether to accept or reject the result (see Fig.~\ref{fig:gameversusverifiable}). The proof \( \Pi \) is a sequence of numbers attached to the output, and it enables the verifier to confirm that the computation was performed correctly without re-executing the task. The DC, or any other participant, can validate the result with minimal overhead.

Given the pair \( (y, \Pi) \), the verifier can immediately decide whether to accept or reject the result, thereby enabling \emph{fast finality}. Moreover, the verifier does not need to trust the prover; correctness is guaranteed by a cryptographic algorithm whose soundness holds independently of node behavior. As a result, the scheme satisfies \emph{resilience in adversarial-majority settings}.

Although verifiable computing is conceptually appealing and provides strong guarantees, it faces two major limitations that hinder its practical deployment:

\begin{itemize}
    \item \textbf{High computational overhead:} Generating validity proofs incurs substantial cost. Even for moderately sized computations, proof generation can be more expensive than the computation itself. A single server is typically limited to computations involving around 100 million gates~\cite{wu2018dizk}, far below the scale required for modern workloads. Despite efforts to reduce this overhead using domain-specific schemes~\cite{feng2021zen, ghodsi2017safetynets, zhao2021veriml, liu2021zkcnn, garg2023experimenting} and parallelization techniques~\cite{wu2018dizk, rahimiJournal}, this remains a fundamental bottleneck. Therefore, verifiable computing does not satisfy the requirement of \emph{low computational overhead}.
    
    \item \textbf{Incompatibility with Approximate Computing:} Most existing verifiable computing frameworks rely on exact arithmetic over finite fields. This makes it difficult to efficiently represent operations such as floating-point arithmetic, quantization, and non-linear activations. As a result, encoding such computations becomes complex and inefficient~\cite{rahimi2025texttt, chen2022interactive, garg2022succinct}, making verifiable computing unsuitable for \emph{approximate computing}.
\end{itemize}

\subsection*{Optimistic Verification}

Another class of solutions, which is inspired by scalability solutions for blockchain like Arbitrum\cite{kalodner2018arbitrum}, is called \emph{Optimistic Verification}. These approaches adopt an \emph{optimistic} approach: the computation is performed off-chain by some external nodes and the result is assumed to be correct unless a dispute is raised.

In this setup, the key players are: (1) a set of \emph{external computing nodes}, responsible for performing the computation task and submitting the result \( y = f(x) \) to the blockchain, and (2) a set of \emph{challengers} (also referred to as \emph{fisherman} or \emph{watchtower nodes}) who monitor submissions for correctness. In the typical (happy) path, the external node executes the task honestly, and the result is accepted without objection. However, if any challenger detects a discrepancy, it can raise an alarm and initiate a dispute process on-chain.

This triggers a fundamental problem: the blockchain must now determine which party is lying: either the external computing node submitted an incorrect result, or the challenger is falsely accusing an honest node. Since both types of behavior are indistinguishable to the blockchain, a verification mechanism is needed to adjudicate the claim.

A naïve solution would be to re-execute the entire computation on-chain. However, this defeats the purpose of outsourcing heavy tasks. To avoid this, optimistic verification frameworks adopt a bisection-style interactive protocol, inspired by the design of optimistic rollups~\cite{kalodner2018arbitrum}.

\begin{figure}[t]
    \centering
    \includegraphics[width=0.48\textwidth]{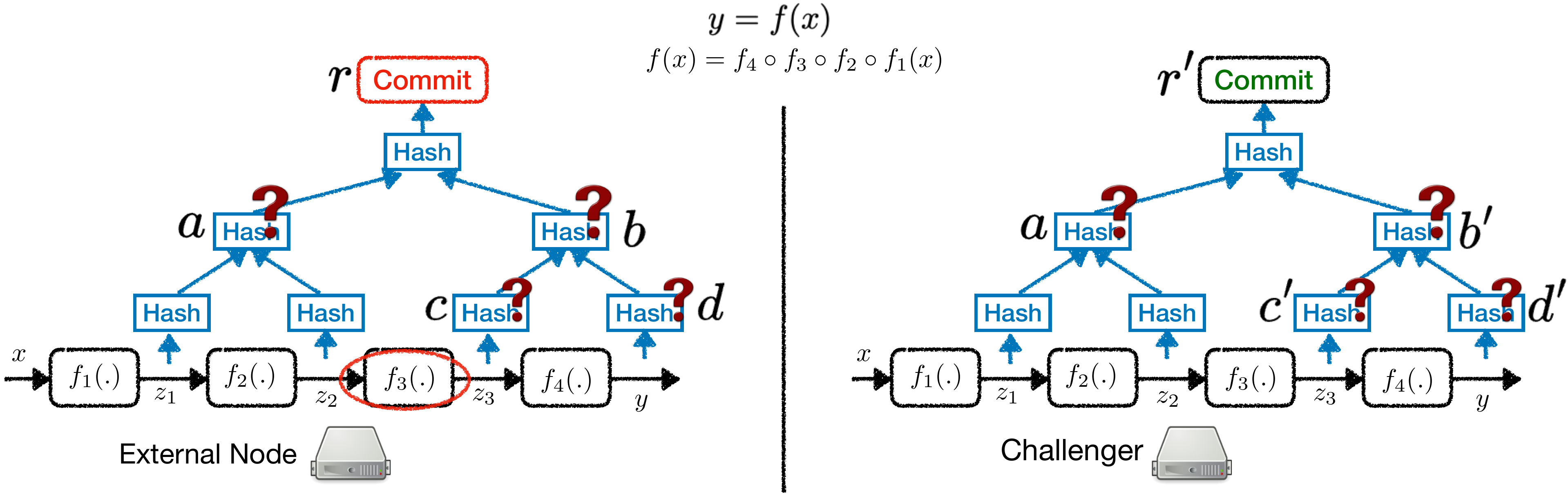}
 \caption{
Illustration of optimistic verification for a layered function \( f(x) = f_4 \circ f_3 \circ f_2 \circ f_1(x) \). The computing node performs the full computation and commits to the intermediate values \( z_1, z_2, z_3, z_4 \) using a binary hash tree: \( a = \mathsf{h}(\mathsf{h}(z_1), \mathsf{h}(z_2)) \), \( b = \mathsf{h}(\mathsf{h}(z_3), \mathsf{h}(z_4)) \), and \( r = \mathsf{h}(a, b) \), which is submitted to the blockchain. If a challenger disputes the result, it independently recomputes the function and submits its own root \( r' \), based on its version of the intermediate values. For example, assume the first point of discrepancy is at \( z_3 \). Since \( r \neq r' \), both parties reveal the corresponding hash preimages. In this case, they agree on \( a \) but disagree on \( b \), prompting the reveal of \( z_3, z_4 \) versus \( z_3', z_4' \). The blockchain then verifies the first differing subfunction, in this case \( f_3 \), by checking whether \( f_3(z_2) = z_3 \), using only on-chain execution of the lightweight function \( f_3 \).
}

    \label{fig:opml}
\end{figure}

More precisely, optimistic verification structures the computation as a composition of lightweight sub-functions. For example, as illustrated in Fig.~\ref{fig:opml}, suppose \( f(x) = f_4 \circ f_3 \circ f_2 \circ f_1(x) \), where each sub-function \( f_i(\cdot) \) is simple and lightweight enough to be executed on-chain, but the full composition \( f(x) \) is too complex or costly to be computed within blockchain constraints.
Let us denote the intermediate values as \( z_1 = f_1(x), z_2 = f_2(z_1), z_3 = f_3(z_2) \), and \( y = z_4 = f_4(z_3) \).

The external node performs the full computation off-chain, obtaining the final output \( y \). In addition, it constructs a tree-like commitment to the intermediate values, i.e., \( z_1, z_2, z_3 \), together with the final result \( z_4 = y \). Let \( \mathsf{h}(\cdot) \) denote a cryptographic hash function. A cryptographic hash function \( \mathsf{h} : \{0,1\}^* \rightarrow \{0,1\}^n \) is a deterministic mapping from inputs of arbitrary length to fixed-length outputs. It is designed to be \emph{collision-resistant}, meaning it is computationally hard to find \( x \neq x' \) such that \( \mathsf{h}(x) = \mathsf{h}(x') \).

The external computing node calculates $ a = \mathsf{h}(\mathsf{h}(z_1), \mathsf{h}(z_2))$, $ b = \mathsf{h}(\mathsf{h}(z_3), \mathsf{h}(z_4))$, and \( r = \mathsf{h}(a, b) \),  
and submits both the result \( y \) and the root hash \( r \) to the blockchain (see Fig.~\ref{fig:opml}). This commitment serves as a compact representation of the computation trace, binding the external computing node to the reported output and its corresponding intermediate states. This structure is known as a Merkle tree.

In the absence of disputes, this submission is accepted. However, if a challenger detects a discrepancy,  
it recomputes the function and constructs its own commitment root \( r' \), based on its intermediate results, using the same structure.  
If \( r \neq r' \), the blockchain initiates an interactive dispute process to identify the first point of disagreement. 

For instance, as depicted in Fig.~\ref{fig:opml}, assume that the first point of discrepancy is in the value of \( z_3 \). The evaluation process on the blockchain proceeds as follows.  
Both parties are asked to reveal the pre-images of their root hashes. Specifically:
\begin{itemize}
    \item The computing node reveals \( a, b \) such that \( \mathsf{h}(a, b) = r \),
    \item The challenger reveals \( a', b' \) such that \( \mathsf{h}(a', b') = r' \).
\end{itemize}

Since \( r \) and \( r' \) differ, there must be a mismatch in one of the branches. In this example, as depicted in Fig.~\ref{fig:opml}, we have \( a = a' \) and \( b \neq b' \), suggesting the difference lies in the subtree rooted at \( b \). The process then continues recursively:
\begin{itemize}
    \item The computing node reveals \( z_3, z_4 \), such that we have \( \mathsf{h}(\mathsf{h}(z_3), \mathsf{h}(z_4)) = b \),
    \item The challenger reveals \( z_3', z_4' \) such that we have \( \mathsf{h}(\mathsf{h}(z_3'), \mathsf{h}(z_4')) = b' \).
\end{itemize}

This recursion continues until a pair of differing values is found at the leaf level. Crucially, the blockchain only needs to verify the \emph{first point of difference}, since all prior steps are agreed upon. As mentioned earlier, since the first discrepancy occurs at \( z_3 \neq z_3' \), the blockchain simply recomputes \( f_3(z_2) \) and compares it to the claimed \( z_3 \), thereby identifying the dishonest party. Since each \( f_i \) is assumed to be lightweight, this verification remains efficient.

This method naturally extends to any number of composition layers, that is, \( f(x) = f_n \circ f_{n-1} \circ \cdots \circ f_1(x) \), for any \( n \in \mathbb{N} \).
Moreover, the approach extends beyond layered computations: any computation structured as a directed acyclic graph (DAG) can be verified using similar techniques~\cite{bhat2023sakshi, conway2024opml}, where bisection is applied over a topological ordering of the DAG’s nodes.

In optimistic verification, the computing nodes are only required to execute the computation and generate a lightweight proof by computing simple hash functions to construct the root $r$. This process is significantly less complex than the  proof generation required in verifiable computing schemes. As a result, optimistic verification satisfies the requirement of \emph{low computational overhead}.

In addition, correctness in optimistic verification is enforced through an interactive dispute process, not through assumptions about the number of honest nodes. As long as a single honest challenger exists to detect and contest incorrect results, the protocol can identify misbehavior and penalize the dishonest party; therefore, optimistic verification satisfies \emph{resilience in adversarial-majority settings}.

While optimistic verification  offers a lightweight verification model, it suffers from two key limitations:

\begin{itemize}
    \item \textbf{Lack of fast finality:} Optimistic verification systems rely on interactive dispute resolution, which involves multiple on-chain steps such as submitting challenges and revealing hash pre-images. These steps depend on timely transaction inclusion and execution, which may be delayed due to network congestion or censorship by adversarial nodes. To mitigate this risk, optimistic verification frameworks typically enforce long fraud-proof windows—often spanning several days—to allow honest parties to respond. Similar delay patterns appear in optimistic rollups, where finality is inherently postponed due to reliance on challenge periods~\cite{kalodner2018arbitrum}. Therefore, optimistic verification fails to provide \emph{fast finality}.
    
    \item \textbf{Incompatibility with approximate computing:} optimistic verification assumes deterministic computation; however, approximate computing introduces non-determinism even among honest nodes. Slight differences in outputs can yield mismatched hash commitments, which are wrongly interpreted as signs of incorrect computation. This undermines dispute resolution and renders optimistic verification protocols ineffective for \emph{approximate computing}.
\end{itemize}

\subsection*{Coded (Redundant) Computing}

\begin{figure}[t]
    \centering
    \includegraphics[width=0.4\textwidth]{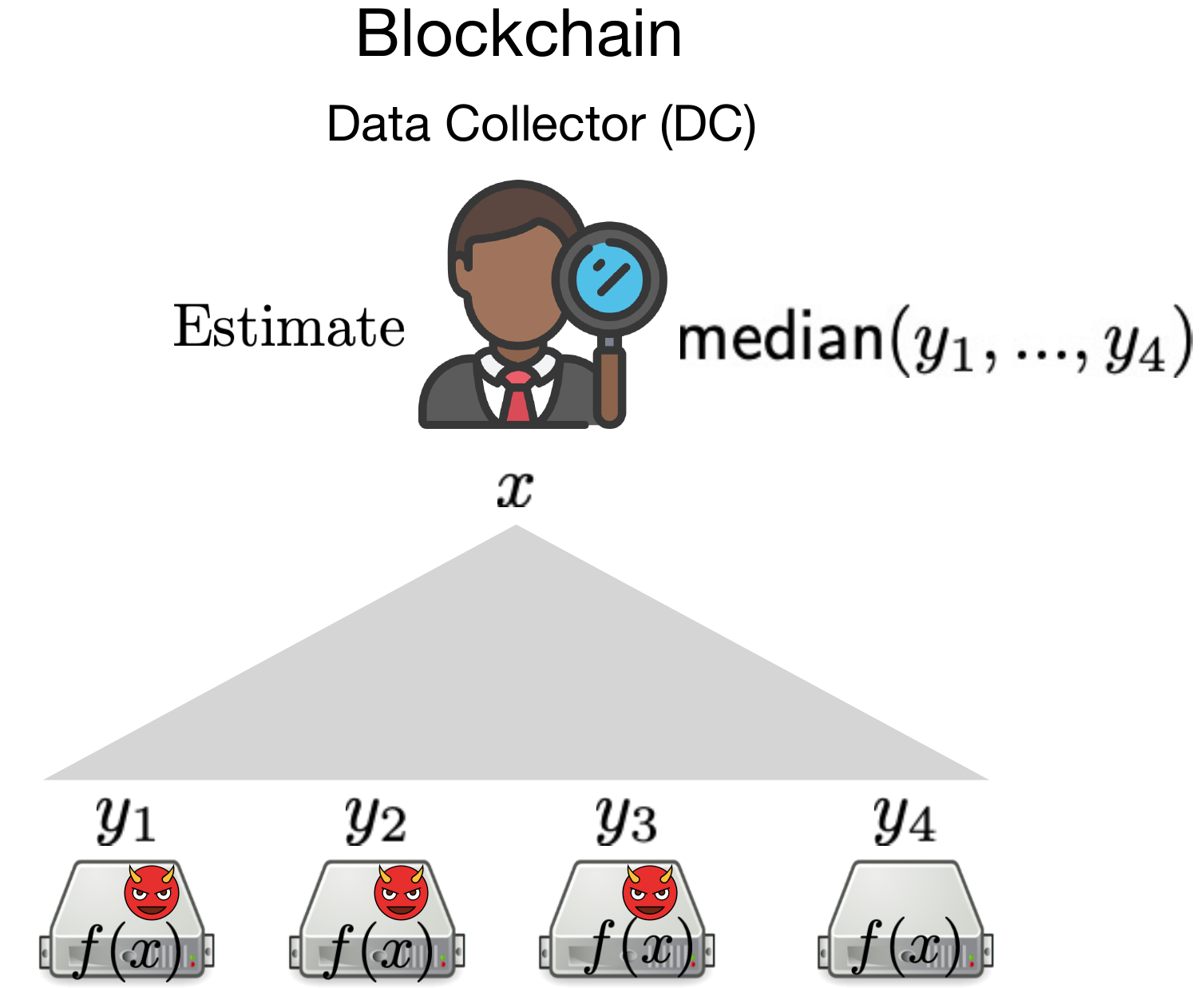}
    \caption{
In coded (redundant) computing, the task is distributed among multiple external nodes using repetition or error-correcting codes. Some of these nodes may be adversarial, but a decoding function (e.g., median or polynomial interpolation) aggregates the results to eliminate errors. This method avoids the need for proof generation and supports approximate computing. It also provides fast confirmation. However, it fundamentally relies on an honest majority assumption.
}
    \label{fig:coded_computing}
\end{figure}

The problem of verifiably outsourcing of the computation motivates  another class of solutions through introducing redundancy via repetition or error-correcting codes \cite{dutta2019optimal, dutta2016short, yu2017polynomial, yu2020entangled}. In this approach, a computational task is assigned to \( N \) external nodes, each of which performs either the same task or a coded version of it. The results are then aggregated and decoded using a reconstruction rule such as majority, median, or algebraic decoding, depending on the coding scheme (see Figure \ref{fig:coded_computing}).

Coding-based solutions offer several favorable properties:

\begin{itemize}
    \item \textbf{Low computational overhead:} Computing nodes are only required to execute the task and return the output, without producing validity proofs. This makes the protocol lightweight and efficient.
    
    \item \textbf{Compatibility with approximate computing:} Real-valued coding schemes are inherently tolerant to minor variations. They support floating-point arithmetic and quantized operations, aligning well with the approximate computing \cite{ moradi2025general, jahani2018codedsketch, roth2020analog}.
    
    \item \textbf{Fast finality:} Once the responses are collected, decoding can be performed immediately. There is no reliance on challenge periods or interactive verification, allowing rapid completion of the protocol.
\end{itemize}

However, this scheme suffers from a critical limitation: it is not resilient against adversarial-majority. Its effectiveness depends on a strong trust assumption. For example, under repetition coding, let \( \mathcal{H} \) denote the set of honest nodes and \( \mathcal{T} \) the set of adversarial ones. Then, correct decoding requires \( |\mathcal{H}| \geq |\mathcal{T}| + 1 \). For a Reed-Solomon \( (K,N) \) code~\cite{SudanBook}, one requires \( |\mathcal{H}| \geq |\mathcal{T}| + K \). Similarly, in coded polynomial computation using Lagrange codes~\cite{yu2019lagrange}, for a polynomial of degree \( d \), correctness requires \( |\mathcal{H}| > |\mathcal{T}| + (K-1)d \). Analog coding schemes~\cite{ moradi2025general, jahani2018codedsketch, roth2020analog} also impose similar trust assumptions. In all cases, error correction is only feasible when the number of honest nodes surpasses the adversarial nodes by a specific margin.

\subsection*{A Fundamental Question}
To summarize, Table~\ref{tab:comparison} compares the approaches discussed so far. This comparison raises a fundamental question: 
\begin{tcolorbox}[colback=gray!10, colframe=black, boxrule=0.5pt]
Can we design a framework that combines the efficiency and flexibility of coding-based methods with resilience to adversarial-majority?
\end{tcolorbox}
More specifically, is it possible to extend coding theory beyond its classical trust assumptions and make it applicable in trust-minimized environments?
These questions motivate the development of the \emph{game of coding} framework~\cite{GOC_Firstpaper, GoDSybil, GOC_unknown}, a fundamentally new approach to computation outsourcing, particularly suited for DeML applications. 
This approach, which is outlined in the next section, incorporates adversarial incentives and game-theoretic reasoning to enable robust computation outsourcing, using coding theory, even under adversarial-majority setting.

\section*{Game of Repetition Coding: Problem formulation}\label{sec:framework}

\begin{figure}[t]
    \centering
\includegraphics[width=0.98\linewidth]{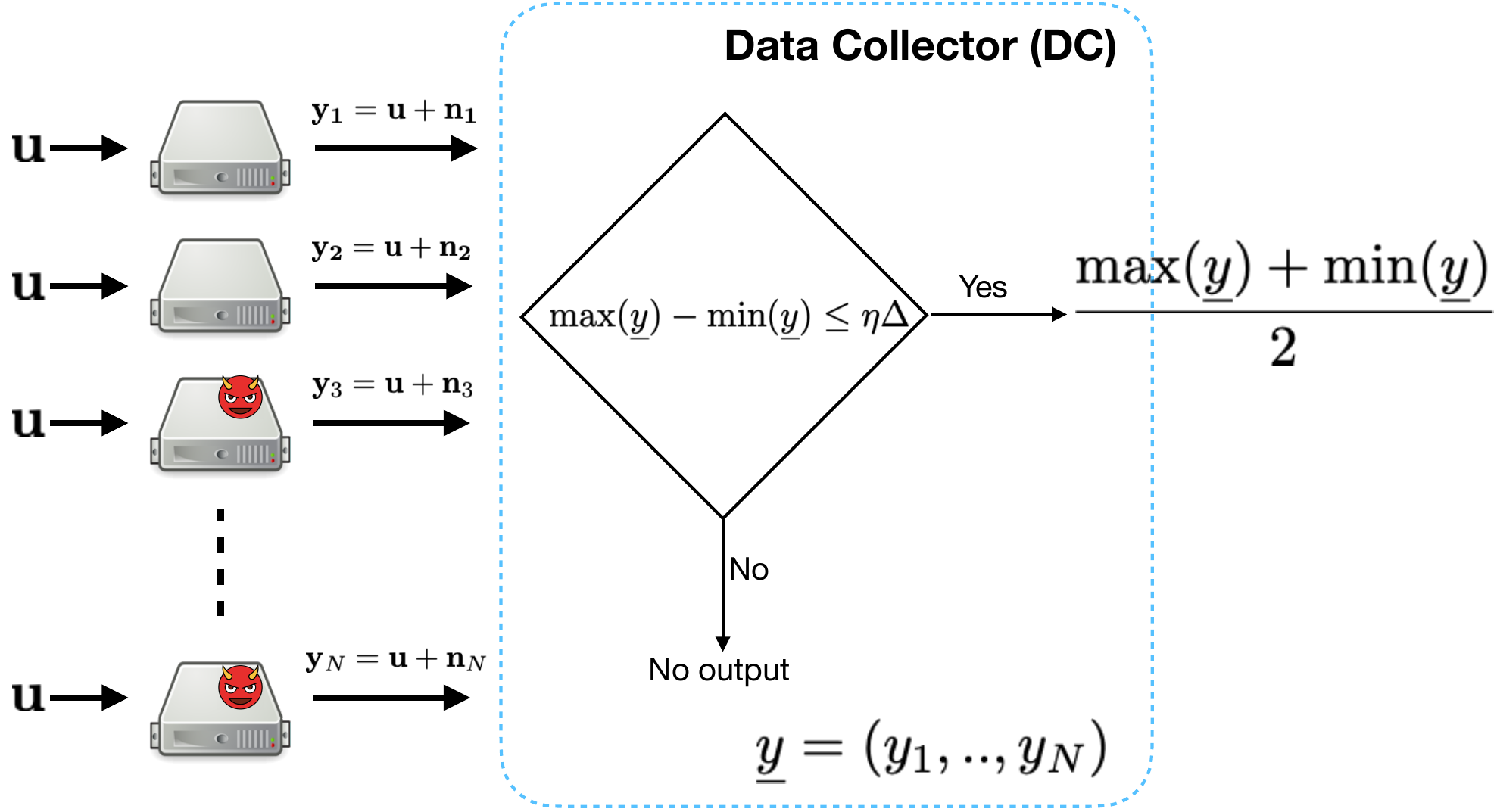}
\caption{An illustration of the game of coding framework. The DC receives responses from multiple computing nodes, which may include both honest and adversarial participants. Each honest node sends a noisy version of its computed value, while adversarial nodes may report arbitrary or strategically manipulated outputs. The DC applies an acceptance rule: it accepts the responses if the range between the maximum and minimum reported values satisfies \( \max(\underline{\mathbf{y}}) - \min(\underline{\mathbf{y}}) \leq \eta \Delta \), where \( \eta \) is the acceptance threshold and \( \Delta \) bounds the honest noise. Upon acceptance, the DC estimates the unknown value \( \mathbf{u} \) by computing the average of the maximum and minimum inputs, i.e., \( \hat{\mathbf{u}} = \frac{\max(\underline{\mathbf{y}}) + \min(\underline{\mathbf{y}})}{2} \). The DC first chooses the threshold \( \eta \), and in response, the adversary selects a noise distribution to maximize its own utility, subject to the DC’s acceptance policy.
}

    \label{fig:model}
\end{figure}
In this section, we review the \emph{game of coding} problem formulation using for basic repetition coding. For the formal  formulation, see\cite{GOC_Firstpaper, GoDSybil, GOC_unknown}.
We consider a system (see Fig.~\ref{fig:model}) composed of a DC and \( N \) external computing nodes. Let \( \mathbf{u} \) be a random variable, uniformly distributed over \( [-M, M] \), where \( M \in \mathbb{R} \). We assume \( \mathbf{u} \) denotes the output of a computation that the DC is unable to perform itself or cannot access directly. To estimate \( \mathbf{u} \), the DC outsources the task to the computing nodes. These nodes execute the delegated computation and return their outputs to the DC. However, not all nodes can be trusted. Some may act honestly, while others may behave adversarially. We denote the set of honest nodes by \( \mathcal{H} \subseteq [N] \), and the set of adversarial nodes by \( \mathcal{T} = [N] \setminus \mathcal{H} \).

Each honest node \( i \in \mathcal{H} \) sends \( \mathbf{y}_i \triangleq \mathbf{u} + \mathbf{n}_i \), where \( \mathbf{n}_i \) has a uniform distribution over the bounded interval \([-\Delta, \Delta]\), for some \( \Delta \in \mathbb{R} \). For simplicity, we assume this uniform distribution in the analysis. For the more general case where \( \mathbf{n}_i \) has a symmetric probability density function (PDF) over the same interval, see \cite{GOC_Firstpaper, GoDSybil}.

 In computational settings, $\bn_h$ may capture rounding noise from fixed-point or floating-point arithmetic, or uncertainty arising from approximate computation methods such as sketching, randomized quantization, random projection, or sampling. These approaches are commonly employed when exact computation is either infeasible or complex.  Throughout, we assume that \( \{ \mathbf{n}_i \}_{i \in \mathcal{H}} \) are independent of $\bu$. 

On the other hand, the adversarial nodes send \( \mathbf{y}_a = \mathbf{u} + \mathbf{n}_a \) to the DC, where \( \{ \mathbf{n}_a \}_{a \in \mathcal{T}} \sim g(\{n_a\}_{a \in \mathcal{T}}) \), for some joint PDF \( g(\cdot) \). The adversary is free to choose the distribution of \( g(\cdot) \) strategically, potentially to mislead the DC. We assume that \( M \) and the distribution of \( \mathbf{n}_h \) are known by all parties, and we have \( \Delta \ll M \).

Upon receiving the vector \( \underline{\mathbf{y}} = (\mathbf{y}_1, \mathbf{y}_2, \ldots, \mathbf{y}_N) \), the DC must decide whether to accept the inputs or reject them. A fundamental question is: under what conditions should acceptance occur? Even in the case where all nodes behave honestly, their reported values \( \mathbf{y}_i \) may differ due to the inherent randomness in the honest noise. However, since each noise term is bounded by \( \Delta \), the maximum difference between the outputs satisfies \( \max(\underline{\mathbf{y}}) - \min(\underline{\mathbf{y}}) \leq 2\Delta \). This insight motivates a natural acceptance criterion in which the DC accepts the inputs if \( \max(\underline{\mathbf{y}}) - \min(\underline{\mathbf{y}}) \leq 2\Delta \).

While the narrow acceptance rule described above may appear reasonable at first glance, it does not necessarily align with the broader interests of the DC. This rule strictly rejects any input set that violates the \( 2\Delta \) threshold, even in cases where the combined inputs allow the DC to form a sufficiently accurate estimate of \( \mathbf{u} \). From a system design perspective, such rigidity may unnecessarily compromise \emph{liveness}. In practice, the adversary might deviate from ideal behavior but still submit values that, when combined with the honest nodes’ outputs, enable the DC to recover \( \mathbf{u} \) with acceptable precision. In such scenarios, it may be beneficial for the DC to tolerate modest deviations in order to preserve system functionality.

\emph{Liveness} is also a critical consideration for the adversary, for at least two key reasons. First, in systems that provide incentives for accepted submissions, the adversary receives a reward only when its inputs are accepted by the DC. Second, the adversary can only influence the final estimate of \( \mathbf{u} \) if the system remains live and the inputs are not rejected. If the system rejects the inputs, the adversary loses any opportunity to affect the outcome. These factors motivate the adversary to behave \emph{rationally} in selecting the joint distribution of \( \{ \mathbf{n}_a \}_{a \in \mathcal{T}} \), optimizing it with respect to its overall utility. On the other hand, the DC may also be incentivized to relax its acceptance criterion to a broader threshold of the form \( \max(\underline{\mathbf{y}}) - \min(\underline{\mathbf{y}}) \leq \eta \Delta \), where \( \eta \in \mathbb{R} \) and \( \eta > 2 \), in order to improve liveness (or the chance of accepting the inputs) while maintaining acceptable estimation quality.

The parameter \( \eta \) controls a fundamental trade-off between liveness and estimation accuracy. At the extreme \( \eta = \infty \), the DC accepts all inputs, ensuring maximal liveness. However, this comes at the cost of accuracy, as the adversary can introduce  unbounded error. On the other hand, setting \( \eta = 2 \) enforces a strict threshold that closely mirrors the case where all nodes are honest, thereby maximizing accuracy. Yet, such a conservative policy exposes the system to denial-of-service attacks. A rational adversary could submit slightly inconsistent values, causing the DC to reject all inputs, even in scenarios where a reasonably accurate estimate of \( \mathbf{u} \) would have been attainable.

This motivation raises two important questions: 
\begin{enumerate}
    \item What is the optimal choice for \( \eta \) for the DC?
    \item  How should the adversary choose the joint distribution of \( g(.) \)?
\end{enumerate}
Interaction between the DC and the adversary can be be captured in 
a game-theoretic formulation,  called \emph{game of coding}.
In the game of coding framework, both the DC and the adversary aim to optimize their respective utility functions, denoted by \( \mathsf{U}_{\DC}(g(.), \eta) \) and \( \mathsf{U}_{\AD}(g(.), \eta) \). The DC selects an acceptance threshold \( \eta \) from the strategy set \( \Lambda_{\DC} = \left\{ \eta ~\big|~ \eta \geq 2 \right\} \), while the adversary chooses a strategy from \( \Lambda_{\AD} = \left\{ g(.) ~\big|~ g(.)~\text{is a valid joint PDF over } \mathbb{R}^{|\mathcal{T}|} \right\} \).

Let \( \mathcal{A}_{\eta} \) denote the acceptance event, defined as 
\begin{align*}
    \mathcal{A}_{\eta} = \left\{ \max(\underline{\by}) - \min(\underline{\by}) \leq \eta \Delta \right\},
\end{align*}
with acceptance probability 
\begin{align*}
    \mathsf{PA}(g(.), \eta) \triangleq \Pr(\mathcal{A}_{\eta}).
\end{align*}
The DC aims to estimate $\bu$ from the observation $\underline{\by}$. To do so, the DC has several estimator options to consider. In this work, we assume the DC employs the following estimator:
\begin{align*}
    \hat{\bu} = \frac{\max(\underline{\by}) + \min(\underline{\by})}{2},
\end{align*}
to recover $\bu$. In the following sections, we will demonstrate that this estimator possesses several appealing properties and  is also closely related to the minimum mean square error (MMSE) estimator.

We define the mean squared error (MSE) conditioned on acceptance as
\[
\mathsf{MSE}(g(.), \eta) = \mathbb{E}[\left( \hat{\bu} - \bu \right)^2 \mid \mathcal{A}_{\eta}].
\]
Since both the mean squared error ($\mathsf{MSE}$) and the probability of acceptance ($\mathsf{PA}$) are critical for the DC and the adversary, the utility functions of the DC and the adversary are naturally defined as follows:
\begin{align}
    \mathsf{U}_{\DC}(g(.), \eta) &\triangleq Q_{\DC}(\mathsf{MSE}, \mathsf{PA}), \nonumber \\
    \mathsf{U}_{\AD}(g(.), \eta) &\triangleq Q_{\AD}(\mathsf{MSE}, \mathsf{PA}), \nonumber
\end{align}
where \( Q_{\DC}: \mathbb{R}^2 \to \mathbb{R} \) is non-increasing in its first argument and non-decreasing in its second, and \( Q_{\AD} \) is strictly increasing in both. This reflects the fact that both the DC and the adversary seek to improve $\mathsf{PA}$, but they have opposing interests with respect to $\mathsf{MSE}$: the DC aims to minimize it, while the adversary benefits from its increase.

As an example, let 
\begin{align*}
    \mathsf{U}_{\DC}(g(.), \eta) &= -3\log(\mathsf{MSE}) + \log(\mathsf{PA}),
    \nonumber \\
    \mathsf{U}_{\AD}(g(.), \eta) &= 2\log(\mathsf{MSE}) + 15\log(\mathsf{PA}). \nonumber
\end{align*}

In this framework, the DC typically corresponds to a program, often referred to as a smart contract, whose strategy is transparently deployed on a blockchain computing platform.
 The adversary observes the strategy chosen by the DC and then selects its own optimal response. This sequential structure aligns with a class of games known as \emph{Stackelberg games}~\cite{von2010market}, where one player, the leader, commits to a strategy first, and the second player, the follower, best responds. In our context, the DC is the leader and the adversary is the follower.

For any value of \( \eta \) committed by the DC, the set of optimal responses of the adversary is defined as  
\begin{align*}
    \mathcal{B}^{\eta}_{\mathsf{AD}} \triangleq \underset{g(.) \in \Lambda_{\mathsf{AD}}}{\arg\max}~ \mathsf{U}_{\mathsf{AD}}\left(g(.), \eta \right).
\end{align*}
All elements of \( \mathcal{B}^{\eta}_{\mathsf{AD}} \) yield the same utility for the adversary, although they may induce different utilities for the DC.  
The DC selects \( \eta \) to maximize its own utility, assuming the adversary responds optimally. To hedge against the worst-case scenario, the DC adopts a pessimistic approach and optimizes for the minimum utility over the adversary’s best-response set:
\begin{align}\label{stackleberg-eqili}
\eta^* = \underset{\eta \in \Lambda_{\mathsf{DC}}}{\arg\max} ~ \underset{g(.) \in \mathcal{B}^{\eta}_{\mathsf{AD}}}{\min} ~ \mathsf{U}_{\mathsf{DC}}\left(g(.), \eta \right).
\end{align}

The main goal is to characterize the Stackelberg equilibrium of the game, that is, (i) to determine the DC’s optimal threshold \( \eta^* \), and (ii) to identify the adversary’s optimal noise distribution, denoted by \( g^*(.) \).

\section*{Finding the Equilibrium }\label{sec:results}

The main challenge in solving \eqref{stackleberg-eqili} lies in its dependence on the utility functions of the adversary and the DC. We make no specific assumptions about these functions, except that the adversary's utility is strictly increasing in both arguments, while the DC's utility is non-increasing in the first and non-decreasing in the second. 
To address this challenge, we introduce an auxiliary optimization problem that is independent of the utility functions. Specifically, for each value \( 0 < \alpha \leq 1 \), we define
\begin{align}\label{C_definition}
     c_{\eta} (\alpha) \triangleq 
    \underset{\gdot \in \Lambda_{\mathsf{AD}}}{\max} ~ \underset{\mathsf{PA} \left( \gdot, \pare \right) \geq \alpha}{\mathsf{MSE}\left(\gdot, \pare \right)}.
\end{align}

It can be shown that for any $\eta$ chosen by the DC, the equilibrium of the game leads to a relationship between $\mathsf{PA}$ and $\mathsf{MSE}$ of the form $\mathsf{MSE} = c_\eta(\mathsf{PA})$ \cite{GOC_Firstpaper, GoDSybil}. This should not be surprising, as for any fixed acceptance probability $\PA$, the adversary seeks to select the noise distribution that maximizes the $\mathsf{MSE}$ (see the definition  \eqref{C_definition}).
Thus, given the function \( c_\eta(\alpha) \), defined in \eqref{C_definition}, the equilibrium of the game can be efficiently computed using the two-dimensional optimization procedure described in Algorithm~\ref{Alg:finding_eta}.
This algorithm takes as input the utility functions \( Q_{\mathsf{AD}}(\cdot, \cdot) \), \( Q_{\mathsf{DC}}(\cdot, \cdot) \), and the curve \( c_\eta(\cdot) \), and outputs the optimal acceptance threshold \( \hat{\eta} \). The correctness of the algorithm is formally established in \cite{GOC_Firstpaper, GoDSybil}.

\begin{algorithm}[t]
\caption{Finding the optimal Acceptance Threshold}
\label{Alg:finding_eta}
\begin{algorithmic}[1]
\State Inputs: $Q_{\mathsf{AD}}(., .), Q_{\mathsf{DC}}(., .)$, $c_{\eta}(.)$ and output: $\hat{\eta}$

\State \textbf{Step 1:} Find  $\mathcal{L}_{\eta} = \underset{0 < \alpha \leq 1 }{\arg\max} ~Q_{\mathsf{AD}}(c_{\eta} (\alpha), \alpha)$
\State \textbf{Step 2:} Output $\hat{\eta} = \underset{\pare \in \Lambda_{\mathsf{DC}}}{\arg\max} ~ \underset{\alpha \in \mathcal{L}_{\eta}}{\min} ~ Q_{\mathsf{DC}} \left(c_{\eta} (\alpha), \alpha\right)$
\end{algorithmic}
\end{algorithm}

Consequently, the remaining step, and the most challenging one,  in computing the equilibrium is to characterize the function \( c_\eta(\alpha) \) defined in \eqref{C_definition}.    

\begin{tcolorbox}[colback=gray!10, colframe=black, boxrule=0.5pt]
Let \( \ell \) denote the number of honest nodes. It can be shown that 
\begin{align}
    c_{\eta} (\alpha) = \frac{h^*_{\eta, \ell}(\alpha)}{4\alpha}, \label{eq:c_of_eta}
\end{align}
where \( h^*_{\eta, \ell}(\alpha) \) is the concave envelop of $ h_{\eta, \ell}(q) \triangleq \nu_{\eta, \ell}(k_{\eta, \ell}^{-1} (q))$, $0 \leq q \leq 1$,  for $\nu_{\eta, \ell}(z) \triangleq \int_{z-\eta\Delta}^{\Delta} (x+z)^2w(x)\,dx$ and $k_{\eta, \ell}(z) \triangleq \int_{z-\eta\Delta}^{\Delta} w(x)  \,dx$, $(\eta-1)\Delta \leq z \leq  (\eta+1)\Delta$, and $w(x) = \frac{\ell}{(2\Delta)^\ell} (\Delta-x)^{(\ell-1)}$.
\end{tcolorbox}
In the process of proving \eqref{eq:c_of_eta} in~\cite{GOC_Firstpaper, GoDSybil}, the optimal adversarial noise distribution \( g^*(\cdot) \) is also characterized, as outlined in Algorithm~\ref{Alg:finding_noise}. It is worth emphasizing that the analysis relies on only minimal assumptions about the utility functions: the adversary's utility is strictly increasing in both arguments, while the DC's utility is non-increasing in the first and non-decreasing in the second. Moreover, although we assume a uniform distribution for the honest noise in this paper for simplicity, the characterization of \( c_{\eta}(\alpha) \) extends to general symmetric noise distributions, as shown in~\cite{GOC_Firstpaper, GoDSybil}. As a result, the formulation of \( c_{\eta}(\alpha) \) applies to a broad class of settings.

A perhaps surprising observation about \eqref{eq:c_of_eta} is that the number of adversarial nodes does not appear in the expression for \( c_\eta(\alpha) \); only the number of honest nodes, \( \ell \), plays a role. In fact, we assume the number of adversarial nodes is just one! Why is this sufficient? This stems from a remarkable property of the game-of-coding framework: the presence of multiple adversarial nodes is effectively equivalent to having a single adversary, due to the system's inherent \emph{Sybil resistance}, a property that will be formally discussed in later sections.

\begin{figure}[t]
    \centering
    \includegraphics[width=0.95\linewidth]{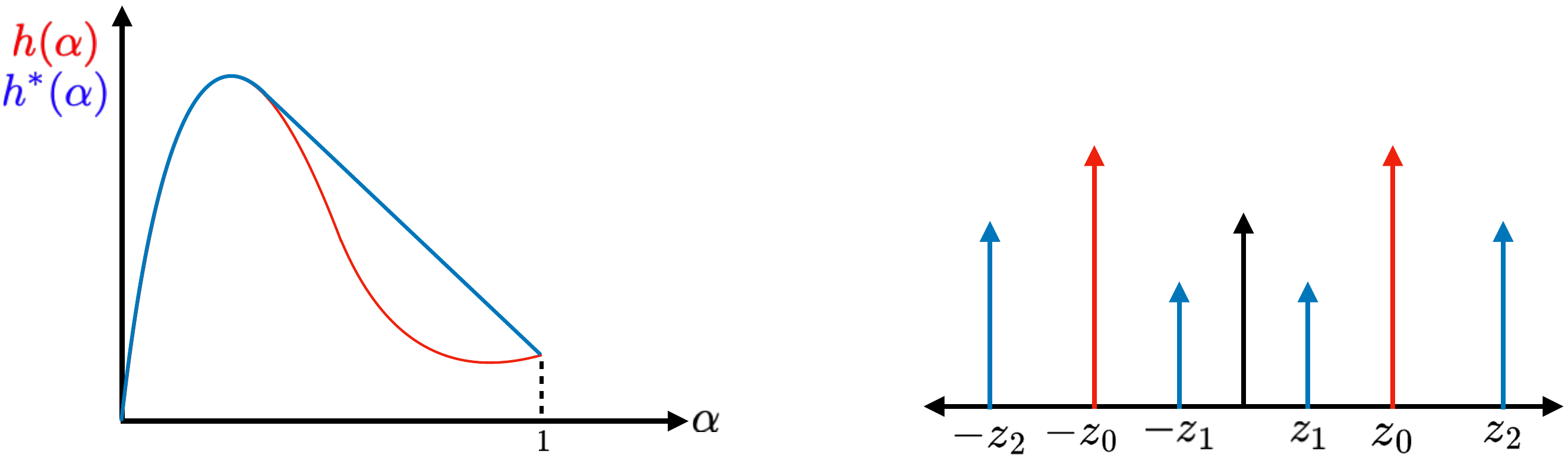}
    \caption{Illustration of the function \( h_{\eta, \ell}(\alpha) \) (red) and its concave envelope \( h^*_{\eta, \ell}(\alpha) \) (blue). The left panel shows the red curve representing the MSE achieved by symmetric single-spike adversarial noise, and the blue curve corresponding to the optimal concave envelope achieved by mixtures of two symmetric spikes. The right panel visualizes the corresponding adversarial noise structures: the red pulses (top) indicate single-spike strategies, and the blue pulses (bottom) illustrate mixtures that interpolate between them to attain the envelope.}
    \label{fig:concave_envelop}
\end{figure}

To prove \eqref{eq:c_of_eta} in~\cite{GOC_Firstpaper, GoDSybil}, the core idea proceeds in two steps. We begin by restricting the adversary’s strategy to a simple class of symmetric noise distributions composed of two Dirac delta functions at symmetric points. Specifically, consider a distribution of the form 
\[
\frac{1}{2}\delta(z - z_0) + \frac{1}{2}\delta(z + z_0),
\]  
which corresponds to the red pulses on the right-hand side of Fig.~\ref{fig:concave_envelop}. For this class of noise, we can determine the optimal value of \( z_0 \) and compute the corresponding mean squared error as
\[
\mathsf{MSE} = \frac{h_{\eta,\ell}(\alpha)}{4\alpha},
\]  
which traces the red curve in the left panel of Fig.~\ref{fig:concave_envelop}.

Next, we show that 
\[
\mathsf{MSE} = \frac{h^*_{\eta,\ell}(\alpha)}{4\alpha},
\]
where $h^*_{\eta,\ell}(\cdot)$ is the concave envelop of the function $h_{\eta,\ell}(\cdot)$, is in fact achievable. To do so, we consider a richer class of noise distributions: symmetric mixtures of two Dirac delta pairs, located at \( \pm z_1 \) and \( \pm z_2 \), with corresponding weights \( \beta_1 \) and \( \beta_2 \). That is, the adversarial noise is distributed as  
\[
\beta_1 \delta(z - z_1) + \beta_1 \delta(z + z_1) + \beta_2 \delta(z - z_2) + \beta_2 \delta(z + z_2),
\]  
which corresponds to the blue pulses in the right panel of Fig.~\ref{fig:concave_envelop}. The optimal values of \( z_1, z_2, \beta_1 \), and \( \beta_2 \) can be computed as shown in Algorithm~\ref{Alg:finding_noise}.

This more expressive distribution allows the adversary to interpolate between two points on the original curve \( h_{\eta,\ell}(\cdot) \), thereby attaining the concave envelope. As a result, the blue curve in the left panel of Fig.~\ref{fig:concave_envelop} becomes achievable. Finally, it can be shown that this envelope is tight: no other noise distribution can yield a better value for \( \mathsf{MSE} \). This completes the proof of \eqref{eq:c_of_eta}.

\begin{algorithm}[t]
\caption{Characterizing the Optimal Noise for Adversary}
\label{Alg:finding_noise}
\begin{algorithmic}[1]
\State Inputs: $Q_{\mathsf{AD}}$, $f_{\bn_h}$, $\eta^*$, $c_{\eta^*}$ and output: $g^*(z)$.

\State \textbf{Step 1:}  Choose $\alpha \in \mathcal{L}_{\eta^*}$, defined in Alg. \ref{Alg:finding_eta}.

\State \textbf{Step 2:}  Consider $h^*_{\eta^*}, h_{\eta^*},$ $ k_{\eta^*}$.
\If {$h^*_{\eta^*}(\alpha) = h_{\eta^*}(\alpha)$}
    \State $g^*(z) = \frac{1}{2}\delta(z+z_1) + \frac{1}{2}\delta(z-z_1)$,  $z_1 \triangleq k^{-1}_{\eta^*}(\alpha)$.
\Else
    \State Find $q_1 < \alpha < q_2$, where $h^*_{\eta^*}(q_1) = h_{\eta^*}(q_1)$, $h^*_{\eta^*}(q_2) = h_{\eta^*}(q_2)$,
    and for all $q_1 \leq q \leq q_2$,  $h^*_{\eta^*}(q) = \frac{h_{\eta^*}(q_2) - h_{\eta^*}(q_1)}{q_2 - q_1} (q - q_1) + h_{\eta^*}(q_1).$ Then $g^*(z) = \beta_1 \delta(z+z_1) +\beta_2 \delta(z+z_2) +\beta_1 \delta(z-z_1) +\beta_2 \delta(z-z_2)$, where $z_1 \triangleq k^{-1}_{\eta^*}(q_1)$, $z_2 \triangleq  k^{-1}_{\eta^*}(q_2)$, $\beta_1 \triangleq  \frac{q_2 -\alpha }{2(q_2 - q_1)}$, $\beta_2 \triangleq  \frac{\alpha - q_1}{2(q_2 - q_1)}$.
 
\EndIf
\end{algorithmic}
\end{algorithm}

\subsection*{Illustrative Examples}
In the following examples, we assume $\ell = 1$, and that the honest node's noise is uniformly distributed over $[-1, 1]$.

\begin{figure}
  \centering
\includegraphics[width=\linewidth]{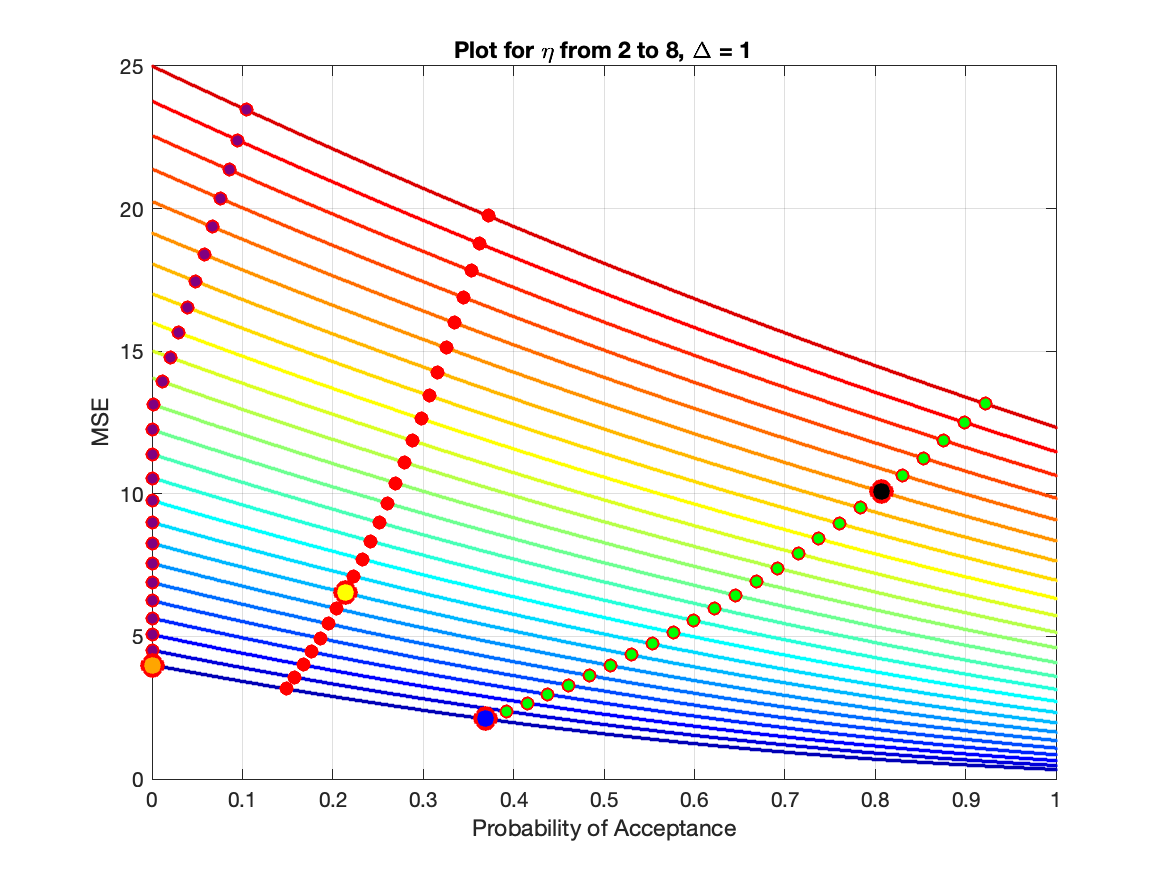}
  \caption{The curves of $c_{\eta}(.)$ for $\eta \in \{2, 2.25, 2.5, \ldots, 8\}$ ($c_{2}(.)$ is the lower-blue curve and $c_{8}(.)$ is the upper-red curve). For each of these $\eta$s,  and the utility functions of Example \ref{first_example_equilibrium}, the green circles represent $\mathcal{L}_{\eta}$.  Here $|\mathcal{L}_{\eta}|=1$ for each $\eta$. The black circle represents the Stackelberg equilibrium for Example \ref{first_example_equilibrium}. 
  The red circles represent $\mathcal{L}_{\eta}$, for the utility functions of Example \ref{second_example_equilibrium}.  Here again $|\mathcal{L}_{\eta}|=1$ for each $\eta$. The yellow circle represents the  Stackelberg equilibrium.
The purple  circles represent $\mathcal{L}_{\eta}$, for the utility functions of Example \ref{example:non_cooperation}.  Here again $|\mathcal{L}_{\eta}|=1$ for each $\eta$. The orange circle represents the  Stackelberg equilibrium.}

  \label{fig:Finding_equilibrium}
\end{figure}

\begin{example}\label{first_example_equilibrium}
Assume that $\mathsf{U}_{\mathsf{AD}}\left( \gdot, \pare \right) = \log  \mathsf{MSE} + \frac{3}{4} \log  \mathsf{PA}$ and 
$\mathsf{U}_{\mathsf{DC}}\left( \gdot, \eta \right) = -\mathsf{MSE} + 25\mathsf{PA}$.
Recall our goals: (i) to find $\eta^*$, the DC’s optimal acceptance threshold, and (ii) to determine $g^*(\cdot)$, the adversary’s optimal strategy.
We characterize the function $c_{\eta}(\cdot)$, shown in Fig.~\ref{fig:Finding_equilibrium}, for $\eta \in \{2, 2.25, 2.5, \ldots, 8\}$. This step is independent of the utility functions. Then, using Algorithm~\ref{Alg:finding_eta}, we compute the adversary's best response. For each $\eta$, we compute the set 
\begin{align*}
    \mathcal{L}_{\eta} = \underset{0 < \alpha \leq 1 }{\arg\max} ~Q_{\mathsf{AD}}(c_{\eta} (\alpha), \alpha).
\end{align*}
In this example, each $\mathcal{L}_{\eta}$ contains a single point, shown as a green circle on the $c_\eta(\cdot)$ curve.
With $\mathcal{L}_{\eta}$ in hand, we compute $\eta^*$ as:
\begin{align}
     \eta^* = \underset{\pare \in \Lambda_{\mathsf{DC}}}{\arg\max} ~ \underset{\alpha \in \mathcal{L}_{\eta}}{\min}  \left ( -c_{\eta}(\alpha) + 25 \alpha \right).
\end{align}
The equilibrium is shown by the black circle in Fig.~\ref{fig:Finding_equilibrium}, corresponding to $\eta^* = 6.75$, with acceptance probability $0.807$ and $\mathsf{MSE} = 10.07$. The adversary’s optimal noise distribution $g^*(\cdot)$ is computed via Algorithm~\ref{Alg:finding_noise}.

For comparison, consider a naive baseline where the DC sets $\eta = 2$, expecting nodes to behave honestly, which implies $\max(\underline{\by}) - \min(\underline{\by}) \leq 2\Delta$. In this case, the adversary responds optimally to its utility, yielding $(\PA, \mathsf{MSE}) = (0.37, 2.10)$, marked with a blue circle in Fig.~\ref{fig:Finding_equilibrium}.
Although this baseline achieves better accuracy ($\mathsf{MSE} = 2.10$ versus $10.07$ at the black circle), the acceptance rate is significantly lower. Hence, the game of coding framework enables the DC to trade off some accuracy in exchange for improved liveness.
\end{example}

\begin{example}\label{second_example_equilibrium}
Assume that $\mathsf{U}_{\mathsf{AD}}\left( \gdot, \pare \right) = \log \mathsf{MSE} + \frac{1}{4} \log \mathsf{PA}$ and 
$\mathsf{U}_{\mathsf{DC}}\left( \gdot, \eta \right) = \mathsf{PA} / \sqrt{\mathsf{MSE}}$.
Following the same procedure as in Example~\ref{first_example_equilibrium}, we compute $\eta^* = 3.75$, with Stackelberg equilibrium $(\PA, \mathsf{MSE}) = (0.214, 6.52)$, shown by the yellow circle in Fig.~\ref{fig:Finding_equilibrium}. For each $\eta$, the corresponding point from $\mathcal{L}_{\eta}$ appears as a red circle on the curve $c_{\eta}(\cdot)$.
\end{example}

\begin{example}\label{example:non_cooperation}
Now suppose $\mathsf{U}_{\mathsf{AD}}\left( \gdot, \pare \right) = \log \mathsf{MSE} + \frac{1}{4} \log(\mathsf{PA} + 0.3)$ and $\mathsf{U}_{\mathsf{DC}}\left( \gdot, \eta \right) = -\mathsf{MSE} + \mathsf{PA}$. 
As in Example~\ref{first_example_equilibrium}, the procedure yields $\eta^* = 2$, with  $(\PA, \mathsf{MSE}) = (0, 4)$, shown as the orange circle in Fig.~\ref{fig:Finding_equilibrium}. This scenario demonstrates a setting where the objectives of the DC and adversary are so misaligned that meaningful cooperation fails. The DC sets a strict threshold $\eta = 2$, and the adversary, in response, selects a noise distribution that results in no accepted outputs at equilibrium.
\end{example}

\section*{Sybil Resistance}

So far, we have reviewed the formulation and equilibrium behavior of the game of coding. In this section, we review its \emph{Sybil resistance} property, originally established in \cite{GoDSybil}.

Sybil resistance refers to the system's ability to remain secure and functional even when the adversary creates multiple fake or duplicate identities (nodes) to manipulate the system. In a Sybil attack, the adversary introduces numerous nodes to overwhelm honest participants, distort decision-making processes, and degrade system performance. For the game of coding framework, ensuring Sybil resistance is critical, otherwise, it would imply that the framework cannot be effectively deployed at scale.

To justify Sybil resistance of the game of coding framework, recall that (see Fig.~\ref{fig:model}) the acceptance condition is  
\begin{align}
    \max(\underline{\mathbf{y}}) - \min(\underline{\mathbf{y}}) \leq \eta \Delta,
\end{align}
and the estimation strategy is
\begin{align}
    \hat{\mathbf{u}} = \frac{\max(\underline{\mathbf{y}}) + \min(\underline{\mathbf{y}})}{2}.
\end{align}

There are two key observations here:

\begin{itemize}
    \item \textbf{For the acceptance, only $\max \underline{\by}$ and $\min \underline{\by}$ matter:} The acceptance rule depends solely on the difference $\max \underline{\by}$ and $\min \underline{\by}$. This means that regardless of how many nodes the adversary controls or how many Sybil identities it introduces, only the $\max \underline{\by}$ and $\min \underline{\by}$ values affect the decision. Any additional nodes reporting intermediate or repeated values have no impact on acceptance.

    \item \textbf{The adversary should not derive both $\max \underline{\by}$ and $\min \underline{\by}$:} The formal proof of this claim can be found in \cite{GoDSybil}. However, this statement can be justified as follows: Recall that the estimator computes the average of the maximum and minimum inputs, i.e., $\max \underline{\by}$ and $\min \underline{\by}$. Thus, if both of these values are chosen by the adversary, then the honest inputs will lie in $[\min \underline{\by}, \max \underline{\by}]$, and closer to estimated value $\frac{\min \underline{\by}+ \max \underline{\by}}{2}$. As a result, the adversary inadvertently  would pull the estimate closer to the true value \( \mathbf{u} \), reducing the estimation error. 
    
    Thus, the optimal strategy for a rational adversary is to control only one of the two $\max \underline{\by}$ and $\min \underline{\by}$ values, rather than injecting multiple Sybil nodes to dominate both. Doing so would not increase the error, and might even reduce it.
\end{itemize}

Motivated by these two key observations, the core idea behind establishing the Sybil resistance of the game of coding, as shown in~\cite{GoDSybil}, is to analyze how the equilibrium behaves under a sequence of progressively simplified adversarial scenarios. The goal is to examine whether an adversary can benefit by replicating itself into multiple identities. If such replication has no effect on the equilibrium behavior, then the system is resistant to Sybil attacks.

\begin{figure}[t]
    \centering
    \includegraphics[width=0.95\linewidth]{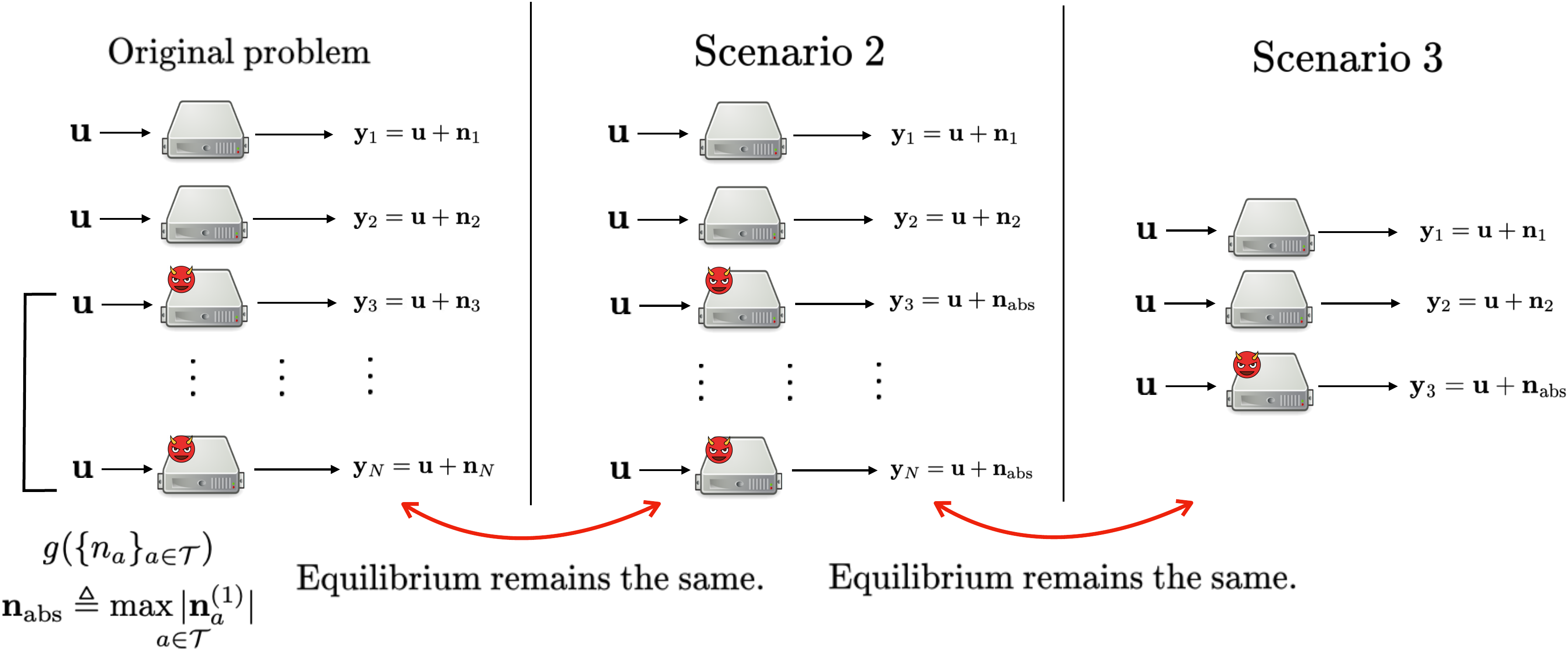}
    \caption{Sybil resistance in the game of coding framework. Left, the original scenario with multiple adversarial nodes; let \( \bn_{\text{abs}} \) be the adversarial noise with the largest absolute value. Middle, all adversarial nodes report the same value \( \bu + \bn_{\text{abs}} \). Right, this setup is reduced to a single adversarial node with the same output. In all cases, the equilibrium remains unchanged.
}
    \label{fig:sybil}
\end{figure}

To demonstrate this,~\cite{GoDSybil} defines and analyzes the following three scenarios, which are also illustrated in Fig.~\ref{fig:sybil}:

\begin{enumerate}
    \item \textbf{Original scenario:} A setting in which both honest and adversarial nodes are present. Among all adversarial nodes, let \( \bn_{\text{abs}} \) denote the noise with the largest absolute deviation from the true value \( \bu \). This extremal adversarial report has the greatest potential influence on the system’s behavior, since the acceptance rule and estimation strategy are based on the maximum and minimum of the reported values.
    
    \item \textbf{Cloned adversary scenario:} The adversary replicates the most extreme behavior across all of its nodes. In this case, every adversarial node reports the same value \( \bu + \bn_{\text{abs}} \). As shown in~\cite{GoDSybil}, this transformation preserves the equilibrium outcome. The maximum and minimum reported values remain the same, so both the acceptance probability and the estimated value are unchanged.
    
    \item \textbf{Single-adversary scenario:} The adversary is reduced to a single node reporting \( \bu + \bn_{\text{abs}} \), while the number of honest nodes remains fixed. Again, it is shown that the system’s equilibrium behavior is preserved. The acceptance decision and the estimation output remain consistent with the previous scenarios.
\end{enumerate}

This sequence of reductions demonstrates that duplicating adversarial behavior across multiple identities does not improve the adversary’s outcome. Whether the adversary reports through many Sybil nodes or through a single node, the system responds identically. Therefore, the game of coding framework satisfies Sybil resistance, making it well-suited for secure and scalable computation outsourcing in adversarial environments.

\section*{The Incomplete Game}

In the initial works on the game of coding~\cite{GOC_Firstpaper, GoDSybil}, it is assumed that the underlying game is \emph{complete}; that is, both DC and the adversary know each other’s utility functions. This assumption is essential for Algorithm~\ref{Alg:finding_eta}, where the DC must evaluate the adversary’s best response, which depends on their utility.

While the adversary can observe the DC’s on-chain strategy, it is less realistic for the DC to know the adversary’s objectives. In practice, this asymmetry gives rise to a \emph{game of incomplete information}, where the DC is uncertain about the adversary’s utility model.

One possible approach is that the DC tries different strategies sequentially. For each choice of \( \eta \), the adversary reacts optimally, and the DC can estimate the corresponding $\mathsf{PA}$ and $\mathsf{MSE}$. Over time, the DC may use these observations to estimate its own utility and converge to the  optimal strategy $\eta^*$. However, this raises a key challenge: while $\mathsf{PA}$ can be estimated, computing $\mathsf{MSE}$ requires knowledge of the true value of \( \mathbf{u} \), which is not accessible to the DC.

This challenge is addressed in~\cite{GOC_unknown}, where the interaction is modeled as a multi-round process. In each round \( t \geq 1 \), the DC selects a threshold \( \eta_t \), and the adversary chooses a response strategy to maximize its utility for that round. The adversary is assumed to be \emph{myopic}, meaning it optimizes its utility myopically based only on the current threshold, without planning across rounds.

After \( T \) rounds, the DC outputs a final decision \( \hat{\eta}_T \). The goal is to design a learning algorithm such that, for any accuracy and confidence parameters \( \lambda, \delta > 0 \), the final strategy satisfies
\[
\Pr \left( \mathsf{U}^* - \mathsf{U}(\hat{\eta}_T) > \lambda \right) < \delta,
\]
where \( \mathsf{U}^* \triangleq \sup_{\eta \in [a, b]} \mathsf{U}(\eta) \) denotes the optimal utility over the feasible range. For simplicity, we use \( \mathsf{U}(\eta) \) to represent the utility of the DC, rather than the more precise notation \( \mathsf{U}_{\mathsf{DC}}(\eta, g(\cdot)) \).

To estimate its own utility at each step, the DC relies on the observable acceptance rate \( \mathsf{PA} \). However, as mentioned earlier, directly estimating \( \mathsf{MSE} \) is challenging, since it requires knowledge of the true value \( \mathbf{u} \). This is resolved using a key structural property shown in~\cite{GOC_Firstpaper}: for any committed threshold \( \eta \) and any equilibrium-inducing adversarial strategy, the pair \( (\mathsf{PA}, \mathsf{MSE}) \) lies on a fixed curve \( \mathsf{MSE} = c_\eta(\mathsf{PA}) \), which is independent of utility functions. This allows the DC to infer \( \mathsf{MSE} \) from the observed \( \mathsf{PA} \).

Building on this,~\cite{GOC_unknown} introduces a sampling-based mechanism (see Algorithm~\ref{Alg:best_eta_not_knowing_Uad}) where the DC sequentially evaluates a grid of candidate thresholds and uses acceptance statistics to estimate utility at each point.

\begin{algorithm}[t]
\caption{Optimal Threshold Selection under Incomplete Information}
\label{Alg:best_eta_not_knowing_Uad}
\begin{algorithmic}[1]
\State \textbf{Inputs:} \( a, b, \delta, \lambda, \ell, L,  c_{\eta}(\cdot), \mathsf{Q}_{\mathsf{DC}}(\cdot) \)
\State \textbf{Output:} \( \hat{\eta} \)

\State Set \( n > (b-a)\max \left\{ \frac{2L}{\lambda}, \frac{1}{d} \right\} \)
\State Set \( k > \frac{8\ell^2}{\lambda^2} \ln\left(\frac{2(n+1)}{\delta}\right) \)
\State Define \( \eta_i = a + \frac{(b-a)(i-1)}{n} \) for \( i \in [n+1] \)

\For{\( r = 1 \) to \( k \)}
    \For{\( i = 1 \) to \( n+1 \)}
        \State Commit to \( \eta_i \), observe whether the response is accepted
    \EndFor
\EndFor

\State For each \( i \), compute \( \hat{\alpha}(\eta_i) = \frac{N(i)}{k} \), where \( N(i) \) is the number of accepted responses
\State Compute \( \hat{\mathsf{U}}(\eta_i) = \mathsf{Q}_{\mathsf{DC}}(\hat{\alpha}(\eta_i), c_{\eta_i}(\hat{\alpha}(\eta_i))) \)
\State Output \( \hat{\eta} = \arg\max_{i} \hat{\mathsf{U}}(\eta_i) \)
\end{algorithmic}
\end{algorithm}

\begin{tcolorbox}
Assume \( \mathsf{U}(\cdot) \) is \( L \)-Lipschitz and \( \mathsf{Q}_\mathsf{DC}(\alpha, c_\eta(\alpha)) \) is \( \ell \)-Lipschitz in \( \alpha \). Then Algorithm~\ref{Alg:best_eta_not_knowing_Uad} guarantees that
\[
\Pr \left( \mathsf{U}^* - \mathsf{U}(\hat{\eta}) > \lambda \right) < \delta.
\]
\end{tcolorbox}

While Algorithm \ref{Alg:best_eta_not_knowing_Uad} provides a robust method for determining the optimal decision region, it can be inefficient in cases where the utility function \( \mathsf{U}(\cdot) \) exhibits significant variations across \([a, b]\). More specifically, if some candidate values of \( \eta \) are unlikely to achieve the maximum utility, they can be dynamically eliminated during the execution of the algorithm. This approach forms the foundation of Algorithm \ref{alg:better_new_alg}.
Algorithm~\ref{alg:better_new_alg} introduces an adaptive mechanism to eliminate suboptimal candidates, similar to the approach used in bandit algorithms.
 Initially, all \( n+1 \) candidates are included. However, at each round \( r \), after committing to the remaining candidates and observing the inputs, the DC calculates the acceptance probability \( \hat{\alpha}(\eta_i, r) = \frac{N(i, r)}{r} \) for \( \{\eta_i\}_{i \in[n+1]\setminus \mathcal{J}}  \), where \( \mathcal{J} \) is the set of eliminated candidates. The DC then computes the utility estimates \( \hat{\mathsf{U}}(\eta_i) \) for the remaining candidates and identifies the current best candidate \( \eta_m \). Using the confidence interval \( \epsilon_r = 2\ell \sqrt{\frac{\ln(4(n+1)/\delta)}{2r}} \) , the DC eliminates any candidate \( \eta_i \) for which \( \hat{\mathsf{U}}(\eta_m) - \hat{\mathsf{U}}(\eta_i) > \epsilon_r \).

In the worst-case scenario, where \( \mathsf{U}(\cdot) \) changes only slightly across \([a, b]\), the DC may not be able to eliminate many candidates, leading to a runtime similar to Algorithm \ref{Alg:best_eta_not_knowing_Uad}. However, in favorable cases, where \( \mathsf{U}(\cdot) \) exhibits sharper variations, the DC can significantly reduce the number of candidates, thus improving runtime efficiency. 
Algorithm \ref{alg:better_new_alg} ensures that the final output achieves a utility that is sufficiently close to the true optimal utility \( \bestu \) while significantly reducing computational overhead.
 The correctness of this algorithm has been shown in \cite{GOC_unknown}.

\begin{algorithm}[t]
\caption{Enhanced Optimal Decision Region Estimation}

\label{alg:better_new_alg}
\begin{algorithmic}[1]
\State \textbf{Inputs:} \( a, b, \delta, \lambda, \ell, L,  c_{\eta}(\cdot) , \mathsf{Q}_{\mathsf{DC}}(\cdot) \).

\State \textbf{Output:} \( \bestetaalgnew \)

\State Choose \( n > (b-a)\max \left\{ \frac{2L}{\lambda}, \frac{1}{d} \right\} \) and \( k > \frac{8\ell^2}{\lambda^2} \ln(\frac{2(n+1)}{\delta}) \).
\State Define \( \eta_i \triangleq a + \frac{(b-a)(i-1)}{n} \), for \( i \in [n+1] \).
\State Initialize \( \mathcal{J} \gets \emptyset \) (the set of eliminated candidates).

\For{\( r = 1, \dots, k \)}

        \State For each \( i \in [n+1] \setminus \mathcal{J} \), commit to \( \eta_i \) and observe the inputs.  Calculate \( N(i, r) \), as the number of accepted inputs for \( \eta_i \), and \( \hat{\alpha}(\eta_i, r) = \frac{N(i, r)}{r} \). Compute \( \hat{\mathsf{U}}(\eta_i) = \mathsf{Q}_{\mathsf{DC}}\big(\hat{\alpha}(\eta_i, r), c_{\eta_i}(\hat{\alpha}(\eta_i, r))\big) \),  \( m = \underset{i \in [n+1] \setminus \mathcal{J}}{\arg\max} \hat{\mathsf{U}}(\eta_i) \),  \( \epsilon_r = 2\ell \sqrt{\frac{\ln(4(n+1)/\delta)}{2r}} \) . If {\( \hat{\mathsf{U}}(\eta_m) - \hat{\mathsf{U}}(\eta_i) > \epsilon_r \)}, add \( i \) to  \( \mathcal{J} \).
\EndFor

\State Find \( m = \underset{i \in [n+1] \setminus \mathcal{J}}{\arg\max} \hat{\mathsf{U}}(\eta_i) \). Output $\bestetaalgnew = \eta_m $.
\end{algorithmic}
\end{algorithm}

\section*{Discussions and Open Problems}
\subsection*{Changing the Estimation Function}

As discussed earlier, the DC uses the estimator 
\begin{align*}
    \hat{\mathbf{u}} = \frac{\max(\underline{\mathbf{y}}) + \min(\underline{\mathbf{y}})}{2}.
\end{align*} 
A natural question is: what if the DC uses a different estimator? For instance, it could adopt the \emph{minimum mean squared error} (MMSE) estimator tailored to the adversary’s noise distribution, defined as
\[
\mathsf{MMSE}(g(.), \eta) = \underset{\est:\mathbb{R}^2 \to \mathbb{R}}{\min}~\mathbb{E}[(\hat{\bu} - \bu)^2 \mid \mathcal{A}_{\eta}].
\]

In~\cite{GOC_Firstpaper}, this question is studied for the case of \( N = 2 \) nodes, one honest and one adversarial. It is shown that, remarkably, for any symmetric noise distribution chosen by the adversary, the estimator \( \frac{\by_1 + \by_2}{2} \) is optimal up to an additive approximation factor of \( \mathcal{O}( \frac{(\eta \Delta)^3}{M} ) \), which vanishes as \( \Delta \ll M \). Building on this, \cite{GOC_Firstpaper} shows that \( \frac{\by_1 + \by_2}{2} \) remains approximately optimal even for general adversarial strategies that achieve equilibrium, regardless of whether the noise distributions are symmetric. 

While the case of \( N = 2 \) nodes has been analyzed, the general case for \( N > 2 \) remains an open and intriguing problem. In fact, it is not even clear whether adopting the MMSE estimator is always beneficial for the DC.

At first glance, using the MMSE estimator may seem optimal for the DC, as it minimizes the estimation error for any noise distribution chosen by the adversary. However, recall that the utility of the DC depends on both the $\mathsf{MSE}$ and the probability of acceptance. Improving the $\mathsf{MSE}$ may unintentionally \emph{reduce} the probability of acceptance, thereby decreasing the overall utility of the DC. While this may seem counterintuitive, it arises from the fact that the adversary’s strategy is influenced by the estimation rule. In a typical equilibrium dynamic, the DC selects both the acceptance threshold \( \eta \) and the estimation function \( \hat{\bu} \), after which the adversary best responds by choosing a noise distribution that maximizes its own utility. This resulting adversarial noise distribution can, in turn, reduce the probability of acceptance and diminish the DC's utility.

Thus, the overall equilibrium is determined by a nested optimization problem: the DC selects its parameters (acceptance threshold and estimator) while anticipating the adversary’s response, and the adversary selects its noise in response to those parameters. In this broader game, it is not evident that the MMSE estimator, which minimizes error, is also the DC’s optimal strategy. Whether MMSE remains optimal in this fully strategic setting is an open question worth exploring.

\subsection*{More Advanced Coding Techniques}

In the initial works on the game of coding~\cite{GOC_Firstpaper, GoDSybil, GOC_unknown}, the underlying scheme is based on \emph{repetition coding}. While this setup enables a clean game-theoretic formulation, an important question arises: \emph{what happens if we replace repetition with more advanced codes?}

To explore this, consider general linear codes, where each node \( n \in [N] \) has access to \( X_n \), and the vector \( \underline{X} = [X_1, \ldots, X_N]^T \) is a coded version of \( \underline{U} \in \mathbb{R}^K \), generated via a matrix \( \mathbf{G} \in \mathbb{R}^{N \times K} \) such that \( \underline{X} = \mathbf{G} \, \underline{U} \). The DC seeks to estimate \( \underline{U} \) based on messages received from all \( N \) nodes. Each honest node \( h \in \mathcal{H} \) sends a noisy version of \( X_h \), while each adversarial node \( a \in \mathcal{A} \) sends a possibly randomized function of \( X_a \).

In this generalized setup, several key questions arise:
\begin{itemize}
    \item \textbf{Acceptance Rule Design:} Unlike repetition, where acceptance is based on pairwise differences, linear codes naturally suggest enforcing approximate \emph{parity checks}. For instance, if \( \mathbf{H} \) is a parity-check matrix for \( \mathbf{G} \), the DC may require \( \|\mathbf{H} \underline{\mathbf{y}} \|_\infty \leq \eta \Delta \) for some threshold \( \eta \).
    
    \item \textbf{Estimation Strategy:} What decoding rule should the DC use to recover \( \mathbf{u} \)? In repetition coding, we considered simple averaging or MMSE. With linear codes, decoding involves inverting \( \mathbf{G} \), which may amplify noise, especially under adversarial manipulation.
    
    \item \textbf{Adversarial Strategy:} The adversary now has more degrees of freedom, not only in crafting corrupt outputs but also in deciding which linear combinations to attack. How does this affect the equilibrium?
    
    \item \textbf{Code Design Criteria:} In classical analog coding, design goals include minimizing noise amplification, bounding condition numbers, or ensuring restricted isometry. In the game of coding, a central open question is whether similar criteria emerge from equilibrium analysis, or a new incentive-aware notions of code quality are needed.
\end{itemize}

These challenges point to a rich extension of the game of coding framework. Moving from repetition to linear codes introduces structure that can be exploited by both the DC and the adversary. This interplay raises several open problems. Understanding these questions is crucial for advancing the theoretical foundation of coding in rational environments, especially when the DC cannot rely on an honest majority.

\subsection*{Proper Pairs of Utility Functions}

As seen in all three examples~\ref{first_example_equilibrium},~\ref{second_example_equilibrium}, and~\ref{example:non_cooperation}, the DC assumes the worst-case scenario by selecting its strategy under the assumption that only one node is honest, regardless of the total number of nodes or the actual number of adversaries.

However, this raises a fundamental question: what happens if the true number of honest nodes is actually larger, say 10 instead of one? At first glance, it seems reasonable to expect that increasing the number of honest nodes would always improve the DC’s utility. Surprisingly, this is not always true.

To see this, consider revisiting the setup of Example~\ref{first_example_equilibrium}. There, the DC calculates \( \eta^* \) based on the most adversarial environment, assuming a minimal honest set. Yet in practice, the number of honest nodes may vary. Since the DC does not know the true numbers of honest and adversarial nodes, it cannot adjust \( \eta \) dynamically. In contrast, the adversary, knowing the full node composition, can adapt its strategy in response to the DC’s committed \( \eta \). This strategic adaptation results in a different pair of values for \( \mathsf{PA} \) and \( \mathsf{MSE} \), distinct from those computed under the worst-case assumption. The final utility of the DC will therefore depend not only on the number of honest nodes but also on the shape of the  curve \( c_\eta(\cdot) \) and the adversary’s utility function.

This effect was studied in~\cite{GoDSybil}, where two illustrative examples were provided. In one case, the DC's utility increased when more honest nodes were present. But in the other, the DC’s utility actually decreased, despite the reduction in adversarial power. This counterintuitive behavior arises because the adversary adapts its noise strategy in response to the new environment, re-optimizing to maximize its own utility. As a result, the net effect on the DC’s utility can be positive or negative, depending on how these interactions unfold.

To formally address this issue, one must verify whether the pair of utility functions is \emph{proper}, meaning that as the number of honest nodes increases, the resulting utility for the DC does not decrease. This ensures that selecting a strategy based on a worst-case assumption remains beneficial or at least safe, even if the actual configuration turns out to be more favorable. A practical approach is to first compute the optimal strategy \( \eta^* \) under the worst-case assumption, then examine whether the resulting utility for the DC improves as the number of honest nodes increases. If so, the utility pair is proper; otherwise, the framework may exhibit non-monotonic behavior.

Finally, this raises an intriguing theoretical question: are there conditions on the utility function of the DC under which the equilibrium pair is guaranteed to be proper, regardless of the adversary’s utility? Characterizing such conditions remains an open and important direction for future research within the game-of-coding framework.

\subsection*{Nonmyopic Adversary}

The analysis in~\cite{GOC_unknown} assumes that the adversary behaves \emph{myopically}, meaning that in each round, the adversary responds  optimally to the current strategy chosen by the DC. This is a natural assumption in settings where the adversary is rewarded based on immediate outcomes. However, one can imagine a \emph{nonmyopic} adversary that anticipates the DC’s learning process and intentionally provides misleading responses in earlier rounds to influence the DC into converging to a wrong long-term strategy. Whether such long-term manipulations can be effective, and how to design DC strategies that are robust to nonmyopic adversaries, remains an open and interesting question in the context of the game of coding framework.

\section*{Conclusion}

Classical coding theory relies on the assumption of an honest majority, where the number of honest nodes exceeds that of adversarial ones by a safe margin. Under this assumption, recovery guarantees are typically established using a worst-case analysis, defending against arbitrary errors regardless of the attacker's strategy. However, in modern incentive-driven systems, this assumption is often too restrictive. These environments reward nodes for accepted contributions and penalize misbehavior, which motivates adversaries to act rationally rather than arbitrarily. As a result, the threat model shifts from worst-case to incentive-aware analysis.

This paper reviews the \emph{game of coding} framework, introduced as a natural extension of classical coding theory that incorporates such rational incentives. Traditional coding guarantees are recovered as a limiting case when adversaries place no value on liveness, resulting in decoding failure at equilibrium. However, for a broader and more realistic class of utility functions, where adversaries do value acceptance, it has been shown that accurate-enough estimation becomes feasible, even in regimes with an adversarial majority~\cite{GOC_Firstpaper}.

The framework models the interaction between a legitimate decoder (DC) and a rational adversary as a two-player Stackelberg game. The DC, acting as the leader, selects an acceptance threshold, while the adversary, as the follower, selects a noise distribution. Each player optimizes their utility, which is defined over two key quantities: the probability of acceptance and the estimation error incurred upon acceptance.

As shown in~\cite{GOC_Firstpaper}, this game-theoretic formulation admits equilibrium strategies under general conditions, and yields improved liveness guarantees in settings where classical coding fails to do so. In~\cite{GoDSybil}, it has further been demonstrated that the framework exhibits \emph{Sybil resistance}—a single honest node can suffice to ensure both estimation accuracy and liveness, even when the adversary splits its influence across multiple identities.

In~\cite{GOC_unknown}, the assumption of complete information was relaxed. While the adversary may observe the DC’s strategy, the reverse is often unrealistic. This gives rise to a game of \emph{incomplete information}, in which the DC lacks knowledge of the adversary’s utility. A learning-based algorithm has been developed to allow the DC to converge to an approximately optimal acceptance strategy, based solely on observable outcomes such as acceptance rate and estimation error.

Taken together, these results provide a principled foundation for incentive-compatible coding in adversarial environments. They open the door for a broader interplay between coding theory, game theory, and real-world computing systems driven by incentives.

\end{document}